\RequirePackage{fix-cm}
\PassOptionsToPackage{table}{xcolor}
\PassOptionsToPackage{hyphens}{url}
\documentclass{fairastra}

\usepackage[T1]{fontenc}
\usepackage[utf8]{inputenc}
\usepackage{lmodern}
\usepackage{newtxtext}
\usepackage{inconsolata}   %
\usepackage{microtype}
\usepackage{array}
\usepackage{tabularx}
\usepackage{amsthm}
\usepackage{amsmath}
\usepackage{amsfonts}
\usepackage{amssymb}
\usepackage{algorithm}
\usepackage{algpseudocodex}
\usepackage{listings}
\usepackage{wrapfig}
\usepackage{float}
\usepackage{enumitem}
\usepackage{url}
\setcitestyle{numbers,square,comma}
\setlist{nosep, leftmargin=*}
\renewcommand{\arraystretch}{0.95}

\theoremstyle{plain}

\theoremstyle{definition}

\theoremstyle{remark}

\newcommand{\mypara}[1]{\vspace{0mm} \noindent \textbf{#1}}

\title{Automating the Design of Embodied Agent Architectures}

\author{\parbox{\linewidth}{\raggedright
Jian Zhou\quad Sihao Lin\quad Jin Li\quad Shuai Fu\quad Gengze Zhou\quad Qi Wu
}}
\renewcommand{\affiliationlist}{{\small\sffamily\bfseries
Australian Institute for Machine Learning, Adelaide University, SA, Australia\par}
{\small\sffamily\{j.zhou, sihao.lin, shuai.fu, gengze.zhou, qi.wu01\}@adelaide.edu.au\quad jinli0410.ai@gmail.com\par}}
\abstract{Embodied agents are typically built as hand-designed compositions of
perception, memory, planning, and action modules. This modularity exposes a
large architectural design space, but current systems still rely on researcher
intuition to choose where information is stored, how observations are processed,
and how model calls are connected. Agent Architecture Search (AAS) automates
such design for text-domain agents, but has not been systematically evaluated
on perceptual embodied agents. We study this transfer. To make embodied architectures searchable, we introduce \textsc{AgentCanvas}
on the executor side, representing embodied agents as editable typed graphs
with simulator execution and episode-level logs. On the optimizer side, we lift AAS from LLM-call loops
to tool-using coding-agent sessions, adapting ADAS and AFlow and introducing
Knowledge Distill Loop (KDLoop). Across four embodied executors (MapGPT, SmartWay, ExploreEQA, and VoxPoser),
search discovers improvements that persist on episodes never queried during search. KDLoop and AFlow independently discover elevation-tolerant observation
processing. KDLoop's implementation also improves success rates over the baseline when transferred to real robots. These improvements are not uniform: some gains remain uncertain or
reverse on held-out episodes, and an apparent gain exploits evaluator-side
information. The experiments further show how rollout noise, local edit basins,
and incomplete use of episode-level evidence constrain search. Together, these results establish the potential of automated architecture
search for embodied agents and characterize the challenges of applying it in
simulated and real-world environments.
Code is available at \url{https://github.com/jianzhou0420/AgentCanvas}.}
\astradata[Project Page]{\url{https://jianzhou0420.github.io/agentcanvas}}
\astradata[GitHub]{\url{https://github.com/jianzhou0420/AgentCanvas}}
\hypersetup{pdftitle={Automating the Design of Embodied Agent Architectures},
  pdfauthor={Jian Zhou, Sihao Lin, Jin Li, Shuai Fu, Gengze Zhou, Qi Wu}}

\makeatletter
\renewcommand{\mymaketitle}{%
  \tcbset{enhanced,frame hidden}
  \tcbset{left=0.5cm}\tcbset{right=0.5cm}\tcbset{top=0.5cm}\tcbset{bottom=0.5cm}
  \tcbset{arc=10pt}\tcbset{colback=astrabg}\tcbset{before skip=0pt}
  \tcbset{grow to left by=1.5pt}\tcbset{grow to right by=1.5pt}
  \begin{tcolorbox}
    \setlength{\parindent}{0cm}
    \setlength{\parskip}{0.5cm}
    {
      \setlength{\parskip}{0cm}
      \raggedright
      \nohyphens{%
      \noindent\begin{minipage}[c]{\dimexpr\linewidth-3.6cm\relax}\raggedright{\setstretch{1.0}\titlelist\par}\end{minipage}\hfill\begin{minipage}[c]{3.2cm}\raggedleft\includegraphics[height=1.35cm,viewport=36 359 1882 1081,clip]{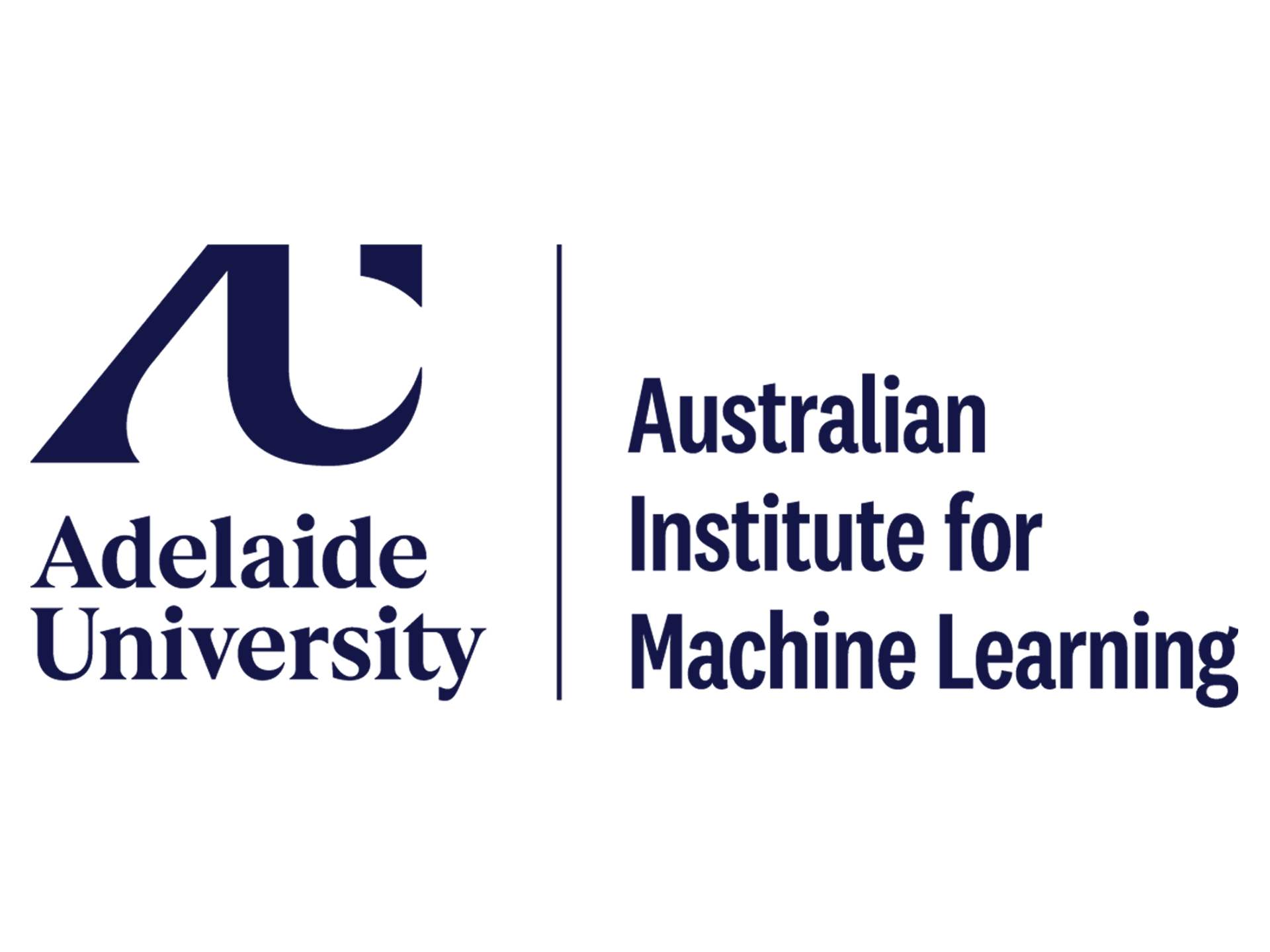}\end{minipage}\par
      }
      \vskip 0.2cm
      \authorlist\par
      \vskip 0.2cm
      \affiliationlist\par
      \contributionlist\par
    }
    \abstractlist\par
    \ifdefempty{\astradatalist}{}{\vskip 0.5cm
    {
      \setlength{\parskip}{0cm}
      \astradatalist\par
    }}
  \end{tcolorbox}
  \tcbset{reset}
  \FloatBarrier
}
\makeatother

\begin{document}
\maketitle

\section{Introduction}
\label{sec:intro}

Embodied agents increasingly act by composing foundation models with perception, mapping, memory, planning, and control modules. This design pattern appears across vision-language navigation~\cite{anderson2018vision,zhou2024navgpt,chen2024mapgpt}, embodied question answering~\cite{das2018embodied,ren2024explore,saxena2024grapheqa}, and language-conditioned manipulation~\cite{liang2023code,huang2023voxposer}, where a system observes the world, builds or queries an internal state, calls language or vision-language models for reasoning, and converts the result into actions.
The advantage of this agentic paradigm is that the architecture is explicit.
Unlike end-to-end policies whose structure is absorbed into weights, these 
systems expose where information flows and which modules can be edited.

This explicit structure also creates a design problem. Each agent fixes choices
about sensor abstractions, map representations, memory state, prompt structure,
planner topology, model placement, and action interfaces. These choices are
usually made by hand for a single benchmark. As foundation models and embodied
tools multiply, the space of plausible architectures grows faster than manual
iteration can cover. The natural question is whether architecture design itself
can be automated.

Agent Architecture Search (AAS) offers one route. In text-domain agents, AAS
uses an optimizer to propose and evaluate alternative LLM-agent workflows,
automating choices that would otherwise be hand-designed
\cite{hu2025automated,zhang2025aflow,shang2025agentsquare,zhang2025multi}.
However, moving AAS to embodied agents is not a direct transfer. Text-domain
AAS operates over cheap, stateless calls and a mature vocabulary of
runnable primitives such as chain-of-thought, debate, or self-refinement.
Embodied agents instead interact with stateful simulators, produce noisy
multi-episode evaluations, and depend on long execution traces involving
observations, actions, tool calls, and intermediate planner outputs. Moreover,
there is no small palette of composable embodied reasoning
primitives: the natural starting points are published systems such as
MapGPT~\cite{chen2024mapgpt}, SmartWay~\cite{shi2025smartway},
ExploreEQA~\cite{ren2024explore}, and VoxPoser~\cite{huang2023voxposer}.

\begin{figure}[t]
\setlength{\belowcaptionskip}{-6pt}
\setlength{\textfloatsep}{8pt}
\centering
\includegraphics[width=\linewidth]{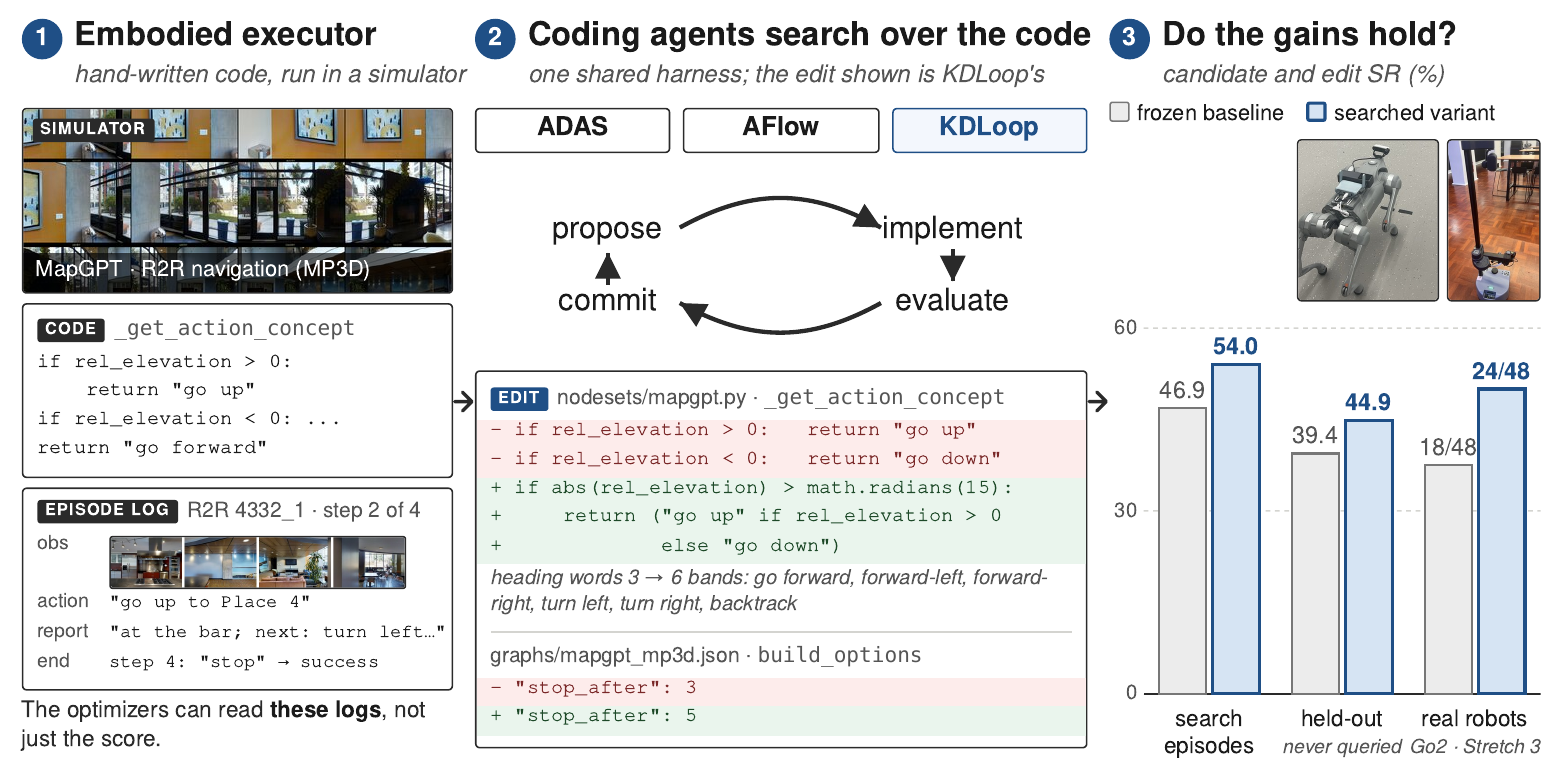}
\vspace{-2.5mm}
\caption{
\textbf{Embodied Agent Architecture Search.}
Coding-agent optimizers (ADAS, AFlow, KDLoop) edit an embodied executor through
one shared harness and score candidates on the search episodes. Selected
candidates are evaluated on held-out episodes never queried during search, and
selected edits are tested on real robots. Shown are KDLoop's MapGPT candidate
and the observation-processing edit tested on robots.
}
\label{fig:teaser}
\end{figure}

We therefore study embodied AAS in a \emph{method-seeded} form. Each search session starts from a published embodied-agent architecture and searches nearby graph-level modifications, rather than assembling an agent from scratch. 
To make this possible, we introduce \textsc{AgentCanvas}, a shared environment
in which coding-agent optimizers edit embodied-agent architectures and evaluate
them through simulator rollouts (Figure~\ref{fig:teaser}). Each executor is
represented as a typed node-and-wire graph, with episode-level logs available
to the optimizer. Within this environment, we implement three search procedures---ADAS~\cite{hu2025automated},
AFlow~\cite{zhang2025aflow}, and \textsc{KDLoop}---to study how architecture
search behaves when applied to embodied agents. \textsc{KDLoop} organizes
search into proposal, critique, experiment, and distillation, with reflection
triggered when progress stalls.

We evaluate the three optimizers across four executors spanning vision-language
navigation, embodied question answering, and language-conditioned manipulation.
Beyond the search results, we evaluate selected candidates on held-out episodes
never queried during search, use add-one ablations to isolate the contribution
of individual edits, and test selected edits on two real-robot platforms.
These evaluations assess whether the gains generalize, which edits account for
them, and whether those edits remain useful on physical robots.

All three MapGPT candidates improve on held-out episodes. For AFlow and
KDLoop, the ablations show that most of the gain comes from observation-processing
edits that include tolerance to elevation noise. The two optimizers discover
these independently and implement them differently. We also verify that KDLoop's implementation improves success
on a Unitree Go2 and a Hello Robot Stretch~3. These results connect the gains
to a specific functional change in observation processing, addressing concerns
about superficial workflow restructuring in text-domain
AAS~\cite{xu2026rethinking}.

The results also show why improvements during search require further
evaluation. ExploreEQA improves on held-out episodes, although both confidence
intervals include zero, and its two tested edits do not improve real-robot
performance. SmartWay--ADAS improves during search but performs worse on
held-out episodes, while a high-scoring SmartWay--AFlow candidate uses
evaluator-side information and is excluded from deployable claims. More
broadly, rollout noise can obscure the effect of edits, search can repeatedly
explore the same local edit basin, and optimizers only partially use
episode-level evidence for credit assignment. Our study therefore identifies
both useful edits and the evaluation and search constraints that must be
addressed when applying AAS to embodied agents.

\vspace{-2mm}

\section{Related Work}
\label{sec:related}
\vspace{-2mm}
\mypara{Embodied Agent Systems.}
Embodied AI has studied agents that connect perception, language, and action in
interactive environments, including vision-language navigation
(VLN)~\cite{anderson2018vision}, embodied question answering
(EQA)~\cite{das2018embodied}, and instruction-following household tasks such as
ALFRED~\cite{shridhar2020alfred}. A growing line of systems composes frozen
LLMs and VLMs with memory, tools, geometric modules, and low-level controllers.
In language-conditioned manipulation, foundation models rank skills, generate
code, or produce geometric constraints for control
(SayCan~\cite{brohan2023can}, Code-as-Policies~\cite{liang2023code},
VoxPoser~\cite{huang2023voxposer}, ReKep~\cite{huang2024rekep}). In
vision-and-language navigation, LLM-based agents reason over textual,
topological, or metric scene representations
(NavGPT~\cite{zhou2024navgpt}, MapGPT~\cite{chen2024mapgpt},
SmartWay~\cite{shi2025smartway}); in embodied question answering, frozen VLMs
are coupled with exploration, memory, and tool use
(Explore-EQA~\cite{ren2024explore}, GraphEQA~\cite{saxena2024grapheqa},
ToolEQA~\cite{zhai2025multi}). These agentic systems differ from embodied
foundation models and end-to-end VLA policies trained or adapted on robot-action
data (PaLM-E~\cite{driess2023palmeembodiedmultimodallanguage},
RT-1~\cite{brohan2023rt1roboticstransformerrealworld},
RT-2~\cite{zitkovich2023rt}, OpenVLA~\cite{kim2024openvla},
$\pi_0$~\cite{black2026pi0visionlanguageactionflowmodel}) in that their
architectures remain explicit. This explicit structure exposes a search space,
but existing systems still rely on hand-designed module choices for each
benchmark.

\mypara{Agent Architecture Search.}
Agent Architecture Search (AAS) automates the design of LLM-agent workflows by
running an external optimizer over candidate architectures at development time.
Our development-time setting complements agents that adapt through
accumulated skills or reflective memory~\cite{wang2023voyager,shinn2023reflexion}. ADAS introduced the
analogy to neural architecture search~\cite{hu2025automated}, and subsequent
work has explored stronger optimizers, structured workflow spaces, and
multi-agent compositions
(AFlow~\cite{zhang2025aflow}, AgentSquare~\cite{shang2025agentsquare},
MaAS~\cite{zhang2025multi}, GPTSwarm~\cite{zhuge2024gptswarm}), as well as
general optimization frameworks and platforms
(Trace~\cite{cheng2024trace}, EvoAgentX~\cite{wang2025evoagentx}). This line is
related to prompt, program, and LM-pipeline optimization methods such as
DSPy~\cite{khattab2023dspycompilingdeclarativelanguage},
OPRO~\cite{yang2024large}, and TextGrad~\cite{yuksekgonul2024textgrad}.
However, AAS treats the agent architecture itself---rather than a prompt,
module, or pipeline parameter---as the object of search. Recent studies show
that text-domain gains can be fragile: tuned single-agent baselines can match
searched workflows~\cite{xu2026rethinking}, and generated workflows may be
unstable under paraphrase~\cite{xu2025robustflow}. Our work tests AAS in a
different regime: perceptual agents acting inside simulators, where
architectures connect perception, memory, planning, and action modules and
evaluations produce noisy multi-episode rollouts.

\mypara{Coding Agents and Graph Substrates.}
Applying AAS to embodied agents requires candidate generation and execution
substrates beyond text-domain workflow search. Prior AAS systems often generate
each candidate with a single LLM completion, sufficient for compact Python
executors but not multi-file embodied-agent code. We therefore use a
coding-agent harness, drawing on systems that edit codebases, run tests, and
iterate (SWE-agent~\cite{yang2024swe},
OpenHands~\cite{wang2025openhands},
Claude Code~\cite{anthropic2024claudecode}). Typed graph substrates such as
LangGraph~\cite{langchain2024langgraph}, AutoGen~\cite{wu2024autogen}, and
Trace~\cite{cheng2024trace} provide related abstractions for composing LLM
calls, but primarily target human-authored workflows or text-domain
optimization. In contrast, embodied AAS needs a graph substrate whose nodes are
bound to stateful simulator execution, action interfaces, and episode-level logs
(\S\ref{sec:substrates}). Unlike repository-level coding-agent benchmarks, we
use off-the-shelf coding agents as embodied-AAS optimizers: each patch changes
an executable agent graph that must remain compatible with simulator APIs,
module input--output interfaces, and diagnostic logging.

\section{Method}
\label{sec:method}

Following prior AAS work~\cite{hu2025automated,zhang2025aflow}, we separate the
task-performing \emph{Executor} from the external \emph{Optimizer} that edits it
across development-time iterations. An \emph{Executor}
is the task-performing agent: a typed graph of perception, memory, planning, and
action modules that runs inside a simulator and returns a task score. An
\emph{Optimizer} is a separate LLM-agent system that edits Executors across
iterations. It never directly acts in the task environment. It proposes graph
edits, triggers evaluation, and uses the resulting code diffs, scores, and logs
to choose subsequent edits.

\subsection{Problem Formulation}
\label{sec:problem}

Let $\mathcal{C}$ be the substrate-supported space of executable
embodied-agent graphs in \S\ref{sec:substrates}.
Each candidate $c\in\mathcal{C}$ receives fitness $f(c)$, measured by success rate
over a multi-episode simulator evaluation, following embodied navigation,
question answering, and manipulation benchmarks
~\cite{anderson2018vision,das2018embodied,shridhar2020alfred,huang2023voxposer}.
Given a seed Executor $c_0$ and budget $B$, an AAS variant generates
a trajectory $\{c_t\}_{t=0}^{B}$ and returns the best evaluated candidate
$c_{t^*}$, where $t^*=\arg\max_{t\le B} f(c_t)$. The transition rule producing $c_{t+1}$ from
the trajectory is the search policy, and is the component that
differs across the three variants in \S\ref{sec:variants}.

Unlike text-domain AAS, we do not search from a small library of generic LLM-call
operators. Embodied agents do not yet have a comparable set of runnable,
task-agnostic primitives: navigation, question answering, and manipulation
systems each contain task-specific perception, state, and action interfaces. We therefore use a \emph{method-seeded} setting (Appendix~\ref{appx:formulation}). Each $c_0$ is a published
embodied-agent method, such as MapGPT~\cite{chen2024mapgpt},
SmartWay~\cite{shi2025smartway}, ExploreEQA~\cite{ren2024explore}, or
VoxPoser~\cite{huang2023voxposer}, and search explores graph-level
modifications around that method.
This setting matches how embodied systems are currently engineered and
lets us ask whether AAS can improve real executors rather than synthetic
workflow sketches.

\subsection{Substrates}
\label{sec:substrates}

\mypara{Executor Substrate: AgentCanvas}
Embodied AAS requires candidates that are both editable by an Optimizer and
runnable inside stateful simulators. We introduce \textsc{AgentCanvas}, a
typed-graph runtime that represents each Executor as a node-and-wire JSON graph
backed by Python node modules. Nodes expose typed input--output ports, and graph
edits---adding edges, swapping nodes, or replacing subgraphs---are applied as
structured patches whose type compatibility is checked before expensive
rollouts.

AgentCanvas also instruments every rollout. Node firings record external inputs
and outputs, while nodes may write internal breadcrumbs such as planner
decisions, prompts, tool calls, or self-reports. These episode-level records
connect candidate edits to downstream behavior and are exposed to all Optimizer
variants through the shared harness, although \S\ref{sec:exp-credit} shows that
access alone does not ensure systematic use.

For scale, simulator-dependent nodes are replicated across workers, while shared
foundation-model nodes remain singleton services that batch requests. This
enables multi-episode evaluation without reloading model weights per worker.
Additional implementation details and runtime contracts are in Appendix~\ref{appx:agentcanvas}.

\mypara{Optimizer Substrate: Coding-agent Harness}
Text-domain AAS often uses a single LLM proposer conditioned on a compact
archive, because candidates are usually small text workflows or compact Python
executors. Embodied executors instead expose multi-file workspaces containing
graph JSON, node code, prompts, evaluation outputs, and episode logs. Useful
proposals must often inspect files, diagnose failed rollouts, and modify both
graph structure and node implementations. A key part of our port is therefore a
repository-level coding-agent substrate that makes existing AAS optimizers
runnable on embodied executors: ADAS and AFlow preserve their proposer and memory
rules, while candidate edits for all variants are implemented through the same
off-the-shelf coding-agent session with file, shell, and tool access
~\cite{yang2024swe,wang2025openhands,anthropic2024claudecode}.

The harness is shared across variants. An outer orchestrator runs the loop: the
variant-specific proposer selects an edit direction, the implementer applies it
to the active graph and code workspace, and the evaluator runs the simulator
suite and records scores and logs (Appendix~\ref{appx:harness}). The variants share implementation tools, execution interfaces, validation,
logging, and file access. Their proposal rules, memory, and allocation of
in-loop evaluations differ. This common infrastructure makes the three search
procedures runnable on the same embodied executors. The study characterizes
their behavior rather than isolating the causal effect of individual optimizer
components or ranking them under matched compute.

This substrate is necessary for embodied AAS, but not sufficient for good
attribution. It records proposals, patches, tool traces, and episode logs; the
Optimizer must still decide which evidence to inspect and how to convert it into
future edits.

\begin{figure}[t]
\centering
\includegraphics[width=\linewidth]{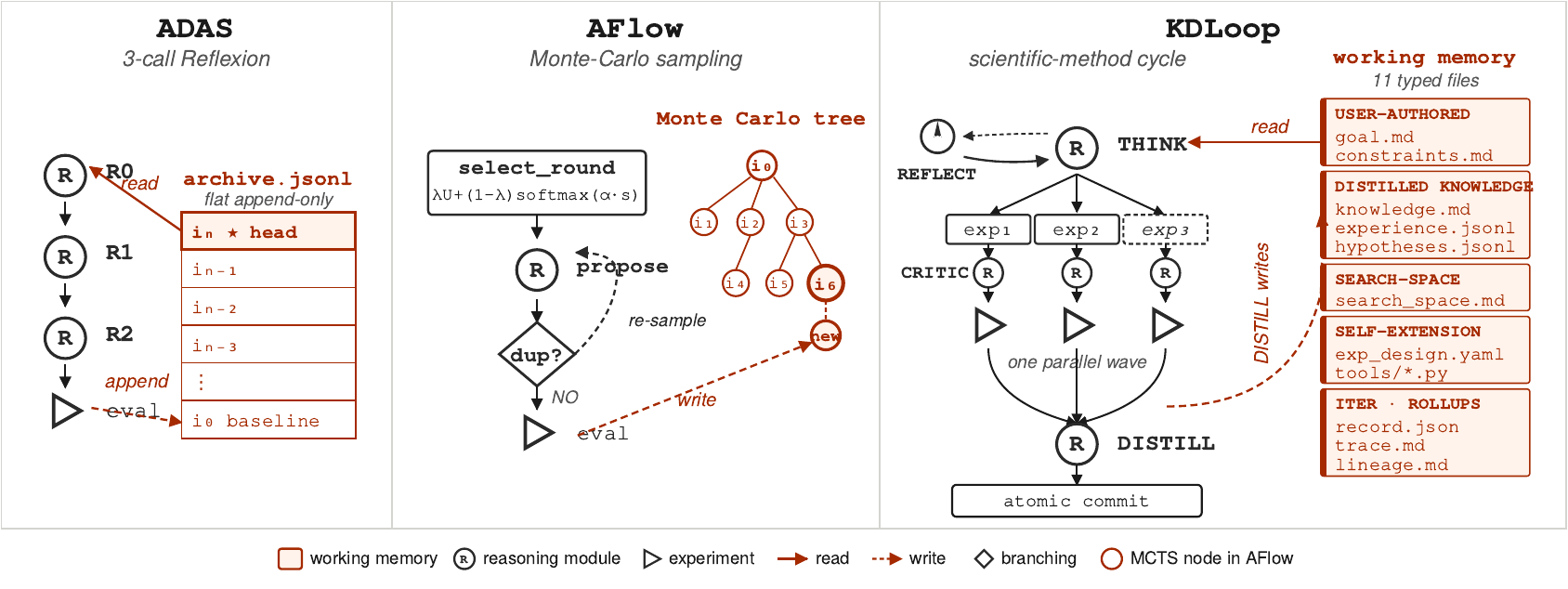}\\[2pt]
\includegraphics[width=\linewidth]{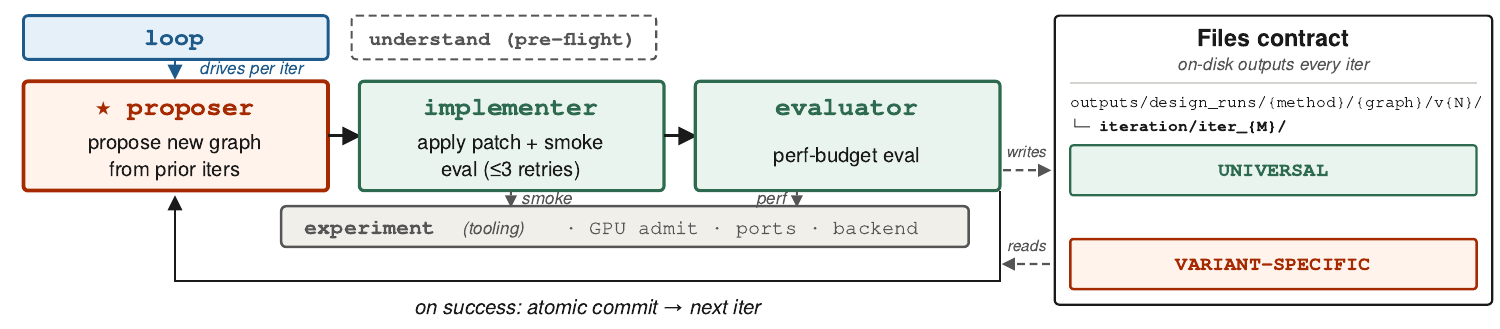}
\caption{\emph{Top}: AAS-method diagrams. Each variant differs in proposer
logic and persistent memory. \emph{Bottom}: the shared coding-agent harness. The
outer loop invokes a proposer, implementer, and evaluator. Proposal and memory rules vary across procedures, as does the allocation of
in-loop evaluations. Implementation tools, simulator execution, validation,
and file access are shared.}
\label{fig:harness}
\end{figure}

\subsection{Variant Implementations}
\label{sec:variants}

We implement ADAS, AFlow, and KDLoop on the common Executor substrate and
coding-agent harness (Figure~\ref{fig:harness}). Each procedure defines how to propose edits, retain
experience, and organize candidate evaluations (Appendix~\ref{appx:impl-details}). ADAS and AFlow adapt existing
text-domain search procedures to embodied executors, while KDLoop organizes
search around explicit hypotheses, experimental outcomes, and coverage of the
editable architecture.

\mypara{ADAS.}
We port ADAS~\cite{hu2025automated} by preserving its Reflexion-style proposal
structure, bootstrap-CI fitness, and flat append-only archive. Each of the original LLM proposal calls is
implemented as an independent tool-augmented sub-agent, so the port retains the
sampling diversity of the original method while giving each proposal access to
the shared codebase and logs.

\mypara{AFlow.}
We port AFlow~\cite{zhang2025aflow} by preserving score-softmax parent
selection, anti-replay memory, and success/failure experience injection. The main adaptation is the search
space: the original AFlow composes over a curated text-workflow operator
library, whereas embodied AAS must edit the seed graph directly. Thus, our port
searches free-form structural changes over typed Executor graphs.

\mypara{KDLoop.}
Knowledge Distill Loop (KDLoop) is designed for the embodied setting, where each iteration produces more
evidence than a scalar score. It runs a four-phase cycle. \textsc{Think} reads
the distilled memory and proposes up to three experiments, each tagged with an
intervention axis such as prompt content, topology, observation pipeline,
state-memory, or model configuration. \textsc{Critic} checks the proposed edits
against previous failed patches and constraints before execution.
\textsc{Experiment} applies and evaluates the selected edit. \textsc{Distill}
writes the outcome back into typed memory, including confirmed findings,
refuted edits, open conjectures, and search-space coverage. A triggered
\textsc{Reflect} phase audits stalled progress, updates coverage, and may mark
parts of the space as exhausted.

Across iterations, KDLoop retains both successful and failed edits as lessons
and tracks unresolved hypotheses separately from completed experiments. The
coverage record tracks which intervention axes have been explored. These
records inform subsequent proposals and the review of stalled progress.
Appendix~\ref{appx:impl-kdloop} details the memory updates and execution cycle.

\section{Experiments}
\label{sec:experiments}

To our knowledge, this is the first systematic evaluation of architecture
search over published perceptual embodied-agent architectures. We evaluate
ADAS, AFlow, and KDLoop across four executors, forming a $3\times4$ matrix.
MapGPT and SmartWay address navigation, ExploreEQA addresses embodied question
answering, and VoxPoser addresses language-conditioned manipulation.

\mypara{Setup.}
Each search starts from the published method's baseline and uses task success
rate (SR) as fitness. Claude Opus~4.7 serves as the search orchestrator, and
executors use \textsc{gpt-5-mini}, except VoxPoser. Candidates are selected on
search episodes and frozen before evaluation on held-out episodes never
queried during search. Baselines and selected candidates are evaluated over
three passes on each available episode set. VoxPoser does not require a
separate held-out split because its LIBERO episodes are near-identical.
Implementation settings, evaluation protocols, and search cost are detailed in
Appendices~\ref{appx:exp-setup}, \ref{app:heldout}, and~\ref{app:search-robustness}.

\begin{table}[t]
\centering
\scriptsize
\setlength{\tabcolsep}{3pt}
\renewcommand{\arraystretch}{1.05}
\caption{
Main results across the $3\times4$ optimizer--executor matrix. SR (\%, mean$\pm$sd
over three passes) of the baseline and the selected candidate on search and
held-out episodes; $\Delta$ in pp.
$^\dagger$ leak-bearing run, excluded from deployable claims and not transferred;
$^\ddagger$ logging-fault diagnosis (\S\ref{sec:exp-credit}); VoxPoser has no
held-out split.
}
\label{tab:headline}
\begin{tabularx}{\linewidth}{@{}l l l >{\raggedright\arraybackslash}X@{}}
\toprule
Executor / Optimizer & Search episodes & Held-out episodes & Surviving change (axis) \\
\midrule
\textbf{MapGPT} & $46.9{\pm}3.1$ & $39.4{\pm}2.1$ & \emph{Baseline} \\
\quad ADAS   & $49.1{\pm}3.2$ ($+2.2$) & $44.4{\pm}1.7$ ($+5.1$) & Step-Back distill + STOP advisor + landmark gate (SM, TO) \\
\quad AFlow  & $54.5{\pm}3.1$ ($+7.6$) & $46.0{\pm}1.4$ ($+6.6$) & Decision rules + history tags + elevation deadband (PC, SM, OP) \\
\quad KDLoop & $54.0{\pm}2.3$ ($+7.1$) & $44.9{\pm}3.5$ ($+5.6$) & Action-concept edit + \texttt{stop\_after}\,$3{\to}5$ (OP, CF) \\
\midrule
\textbf{ExploreEQA} & $43.0{\pm}1.7$ & $44.3{\pm}0.6$ & \emph{Baseline} \\
\quad ADAS   & --- & --- & No iteration beat the baseline \\
\quad AFlow  & $47.7{\pm}2.1$ ($+4.7$) & $48.3{\pm}0.6$ ($+4.0$) & Single-letter VLM prompt (PC) \\
\quad KDLoop & $46.0{\pm}1.0$ ($+3.0$) & $47.7{\pm}1.5$ ($+3.3$) & Answer prior + sentinel filter/prompt + letter-max (SM, OP, PC) \\
\midrule
\textbf{SmartWay} & $29.7{\pm}2.1$ & $27.7{\pm}2.5$ & \emph{Baseline} \\
\quad ADAS            & $33.7{\pm}2.5$ ($+4.0$) & $18.7{\pm}2.3$ ($-9.0$) & 3-sample vote + arrival critic (MC, TO) \\
\quad AFlow$^\dagger$ & $38.7{\pm}5.9$ ($+9.0$) & --- & Evaluator-wired progress tracker (\textbf{leak}) \\
\quad KDLoop          & $31.0{\pm}4.6$ ($+1.3$) & $28.0{\pm}2.6$ ($+0.3$) & STOP-lexicon history accumulator (SM) \\
\midrule
\textbf{VoxPoser} & $\phantom{0}9.0{\pm}0.0$ & --- & \emph{Baseline} \\
\quad ADAS             & $12.5{\pm}0.0$ ($+3.5$) & --- & Composer rewrites + execution-loop tuning (MC, AS) \\
\quad AFlow            & $12.2{\pm}0.6$ ($+3.2$) & --- & \textsc{gpt-4o} composer + retry loop + workspace bounds (MC, CF, OP) \\
\quad KDLoop$^\ddagger$ & --- & --- & Logging fault diagnosed below graph layer; no mutation \\
\bottomrule
\end{tabularx}

\vspace{2pt}
{\scriptsize
PC prompt; TO topology; CF control-flow; OP observation; SM state-memory;
MC model-config; AS action-space.
}
\end{table}

\subsection{Main Results: Does AAS Improve Embodied Executors?}
\label{sec:does-aas-help}

Table~\ref{tab:headline} compares baseline and selected-candidate success
rates on search and held-out episodes, showing whether gains found during
search generalize to unqueried episodes. MapGPT
improves significantly by $+5.1/{+}6.6/{+}5.6$\,pp on held-out episodes for ADAS/AFlow/KDLoop;
\S\ref{sec:exp-ablation} isolates the responsible edits. ExploreEQA also improves on held-out episodes, with gains of $+4.0$ and
$+3.3$\,pp for AFlow and KDLoop, although both confidence intervals include
zero. SmartWay--ADAS instead shows a statistically significant reversal from
$+4.0$\,pp on search episodes to $-9.0$\,pp on held-out episodes, while
SmartWay--KDLoop gains only $+0.3$\,pp held-out, with a confidence interval
that includes zero.

SmartWay--AFlow uses evaluator-side information, so this
leak-bearing candidate is excluded from deployable claims and held-out
evaluation. VoxPoser has no held-out split because its LIBERO episodes are
near-identical.

No optimizer dominates. ADAS tends to concentrate on topology and model
configuration, AFlow often edits the observation pipeline, and KDLoop spreads
edits across observation, prompt, control, and state-memory axes. The two cells
without an SR claim are also informative: ADAS on ExploreEQA produces no
candidate that beats the baseline, while KDLoop diagnoses the shared VoxPoser
logging fault below the graph layer and produces no candidate.

Representative search trajectories and the complete $12$-cell grid are shown in
Appendix~\ref{app:trajectories} (Fig.~\ref{fig:trajectories}).

\subsection{Ablation Study: Which Edits Carry the Gains?}
\label{sec:exp-ablation}

To isolate each edit's contribution, we add it alone to the frozen baseline and
evaluate it over three passes on the same held-out split
(\S\ref{sec:does-aas-help}, Table~\ref{tab:ablation}). On MapGPT, KDLoop and
AFlow independently find observation-processing edits with the largest gains.
KDLoop's action-concept edit combines elevation tolerance with revised heading
descriptions, while AFlow introduces an elevation deadband. These edits add
$+5.7$ and $+5.2$\,pp, close to the full candidates' $+5.6/{+}6.6$\,pp. The other edits
contribute $-3.6$ to $+0.3$\,pp. On SmartWay--ADAS, the three-sample vote
accounts for the reversal ($-7.0$\,pp), while the arrival critic adds
$+1.0$\,pp. ExploreEQA--KDLoop has no dominant edit: the answer prior and
sentinel prompt add $+2.0$ and $+1.7$\,pp, and the sentinel filter and
letter-max contribute $+0.7$ and $-0.6$\,pp. The ablation therefore
distinguishes functional edits from those selected alongside them, a distinction
the optimizers' own credit assignment did not make (\S\ref{sec:exp-credit}).

\begin{table}[t]
\centering
\scriptsize
\setlength{\tabcolsep}{2.5pt}
\renewcommand{\arraystretch}{1.1}
\caption{Add-one ablation on held-out episodes. Each row adds one surviving
edit to the frozen baseline; SR mean$\pm$sd (\%) over three passes, $\Delta$ vs.\
the baseline. K denotes KDLoop and A denotes AFlow.}
\label{tab:ablation}
\begin{tabular}{@{}lr@{\hskip 9pt}lr@{\hskip 9pt}lr@{}}
\toprule
\multicolumn{2}{@{}l}{MapGPT} &
\multicolumn{2}{l}{SmartWay (ADAS)} &
\multicolumn{2}{l@{}}{ExploreEQA (KDLoop)} \\
\midrule
\emph{baseline}           & $39.4{\pm}2.1$          & \emph{baseline}    & $27.7{\pm}2.5$          & \emph{baseline}    & $44.3{\pm}0.6$ \\
$+$stop gate (K)          & $39.7{\pm}2.1$ ($+0.3$) & $+$3-sample vote   & $20.7{\pm}2.5$ ($-7.0$) & $+$answer prior    & $46.3{\pm}1.2$ ($+2.0$) \\
$+$action concept (K)       & $45.1{\pm}3.3$ ($+5.7$) & $+$arrival critic  & $28.7{\pm}3.1$ ($+1.0$) & $+$letter-max      & $43.7{\pm}0.6$ ($-0.6$) \\
$+$elevation deadband (A) & $44.6{\pm}0.7$ ($+5.2$) &                    &                         & $+$sentinel filter & $45.0{\pm}1.7$ ($+0.7$) \\
$+$decision rules (A)     & $39.2{\pm}1.9$ ($-0.2$) &                    &                         & $+$sentinel prompt & $46.0{\pm}3.0$ ($+1.7$) \\
$+$history tags (A)       & $35.8{\pm}0.7$ ($-3.6$) &                    &                         &                    & \\
\bottomrule
\end{tabular}
\end{table}

\subsection{Real-Robot Evaluation: Do the Edits Transfer?}
\label{sec:exp-real}

\begin{figure}[t]
\centering
\newlength{\rpw}\setlength{\rpw}{0.36\textwidth}
\newlength{\ringw}\setlength{\ringw}{0.78\rpw}
\newlength{\ringh}\setlength{\ringh}{0.4968\ringw}
\newcommand{\vrow}[2]{\includegraphics[width=\ringw]{#1}\hfill\includegraphics[height=\ringh]{#2}}
\begin{minipage}[c]{\rpw}\centering
  \vrow{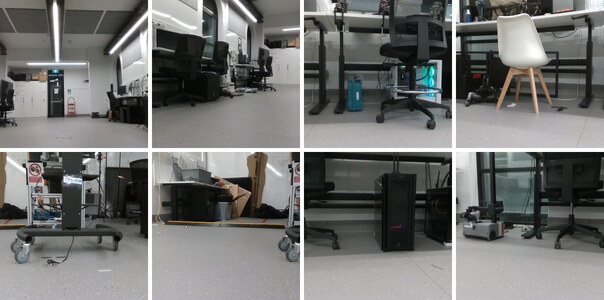}{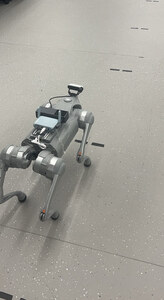}\\[2pt]
  \vrow{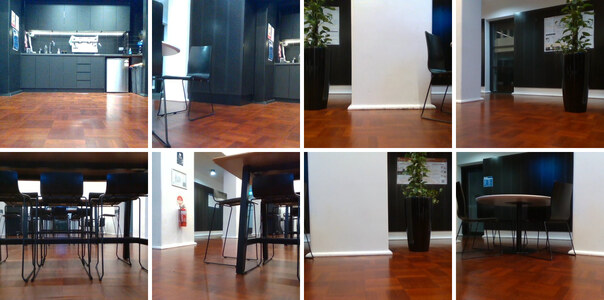}{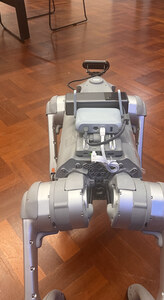}
\end{minipage}\hfill
\begin{minipage}[c]{\rpw}\centering
  \vrow{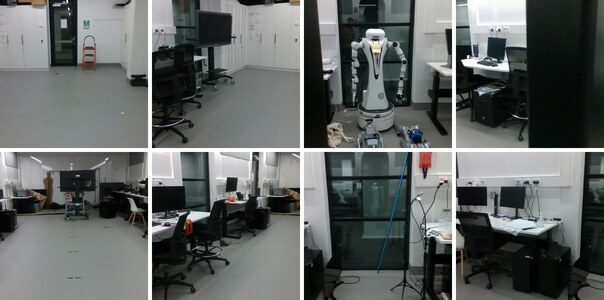}{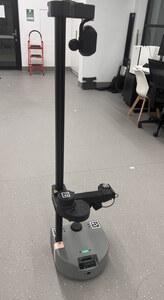}\\[2pt]
  \vrow{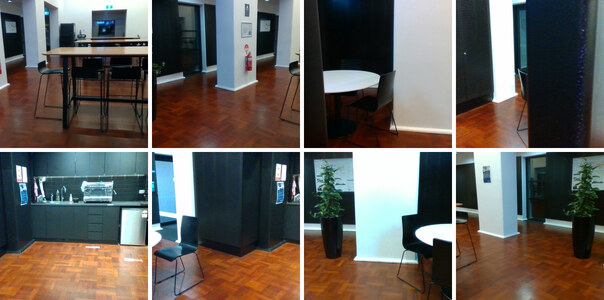}{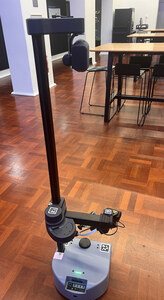}
\end{minipage}\hfill
\begin{minipage}[c]{0.24\textwidth}\centering
  {\scriptsize\setlength{\tabcolsep}{3pt}
  \begin{tabular}{@{}lr@{}}
  \toprule                   & succ./48 \\ \midrule
  MapGPT                     & 18 \\
  \;$+$action concept      & 24 ($+$6) \\
  ExploreEQA                 & 20 \\
  \;$+$answer prior          & 19 ($-$1) \\
  \;$+$sentinel prompt       & 20 (0) \\ \bottomrule
  \end{tabular}}
\end{minipage}
\caption{Real-robot transfer. Left: Unitree Go2; middle: Hello Robot
Stretch~3; each with eight of the twelve panorama views it captured in the
laboratory floor (top row) and the dining space (bottom row). Right: successes
out of 48 episodes per arm, comparing each frozen baseline with versions that
add one edit.}
\label{fig:robot}
\end{figure}

We evaluate five arms on a Unitree Go2 and a Hello Robot Stretch~3 in a
laboratory and a dining space (Figure~\ref{fig:robot}). Each arm runs 48
episodes, 12 per robot and venue pair, using matched tasks within each
executor. The arms are the two frozen baselines, MapGPT with KDLoop's
action-concept edit, and ExploreEQA with either the answer prior or sentinel
prompt, totaling 240 episodes. MapGPT improves from 18/48 to 24/48 successes
(Go2 $5{\to}6$, Stretch $13{\to}18$). ExploreEQA scores 20/48, 19/48, and
20/48 for the baseline, answer prior, and sentinel prompt. These single-run
results provide directional evidence of transfer for the MapGPT edit.

\subsection{Evaluation Noise}
\label{sec:exp-noise}

Within a search run, the optimizer uses candidate evaluations to select edits
and guide subsequent proposals. Rollout noise makes these scores uncertain,
so a high score can reflect a favorable pass rather than a better architecture.
Selecting the best iteration from a sequence of noisy estimates can therefore
favor a lucky candidate and direct further search toward its edits.
The high-variance SmartWay cells illustrate the scale of this noise.
AFlow's selected SmartWay design reruns at $43/32/41\%$, giving
$38.7{\pm}5.9\%$ SR. KDLoop shows the same variance pattern at a lower mean,
$32/35/26\%$, giving $31.0{\pm}4.6\%$ SR. These reruns show that the same
candidate can receive substantially different scores on the same episode set,
making small improvements difficult to distinguish from rollout variability
during search.

Lower-variance cells behave differently. KDLoop on MapGPT
($54.0{\pm}2.3$ vs.\ $46.9{\pm}3.1$) and the two ExploreEQA improvements
($47.7{\pm}2.1$ for AFlow and $46.0{\pm}1.0$ for KDLoop, from
$43.0{\pm}1.7$) exhibit smaller rerun fluctuations than the SmartWay candidates.
The resulting lesson is methodological rather than tied to a particular
executor. Search must account for uncertainty in candidate scores when deciding
which edits to retain and where to explore next. Repeated evaluation during
search can help distinguish consistent improvements from favorable individual
passes, but consumes rollout budget that could otherwise evaluate new edits.
Embodied AAS therefore faces a tradeoff between estimating candidate quality
more reliably and exploring more candidates.

\begin{table}[t]
\centering
\scriptsize
\setlength{\tabcolsep}{3.2pt}
\renewcommand{\arraystretch}{1.03}
\caption{Per-cell attempted intervention-axis counts. Bold marks the repeated
edit family analyzed in \S\ref{sec:local-optima}. Counts include KDLoop--VoxPoser
attempts before the injected logging fault was surfaced.}
\label{tab:diversity}
\begin{tabular}{@{}ll cccccccc@{}}
\toprule
Optimizer & Executor & PC & TO & CF & OP & SM & MC & AS & NP \\
\midrule
\multirow{4}{*}{ADAS}
 & MapGPT     & 1  & \textbf{10} & 3 & 4 & 1 & 1 & -- & -- \\
 & ExploreEQA & -- & -- & 1 & \textbf{9} & -- & -- & -- & \textbf{10} \\
 & SmartWay   & 1  & 7  & 1 & 1 & -- & \textbf{9} & -- & 1 \\
 & VoxPoser   & -- & -- & 6 & -- & -- & \textbf{10} & 3 & 1 \\
\midrule
\multirow{4}{*}{AFlow}
 & MapGPT     & 4 & 2 & 6 & 5 & 2 & -- & 1 & -- \\
 & ExploreEQA & 5 & -- & 2 & \textbf{12} & -- & -- & -- & -- \\
 & SmartWay   & 3 & 2 & 2 & 7 & 1 & 4 & -- & -- \\
 & VoxPoser   & 1 & -- & 4 & 1 & -- & \textbf{8} & 5 & -- \\
\midrule
\multirow{4}{*}{KDLoop}
 & MapGPT     & 2  & 2  & 2 & 3 & -- & -- & -- & -- \\
 & ExploreEQA & 3  & 1  & 1 & 1 & 3  & -- & -- & 1 \\
 & SmartWay   & 2  & 2  & -- & 2 & 3 & -- & 1 & 1 \\
 & VoxPoser   & -- & -- & 4 & 2 & -- & 1 & -- & 1 \\
\bottomrule
\end{tabular}

\vspace{2pt}
{\scriptsize
PC prompt; TO topology; CF control-flow; OP observation; SM state-memory;
MC model-config; AS action-space; NP no-change probe.
}
\end{table}
\subsection{Local Optima in Search Dynamics}
\label{sec:local-optima}

The recorded search trajectories show repeated exploration within local edit
families. Once ADAS or AFlow finds a
high-scoring modification pathway, later proposals often remain near the same
mechanism. Although both ports edit executable graphs, neither records which
intervention axes have already been explored, whether an axis is saturated, or
whether the editable space has been broadly covered. KDLoop explicitly tags
proposals by intervention axis and uses typed history to redirect search after
stalls.

Two observations illustrate this concentration. In Table~\ref{tab:diversity},
the mean per-cell share of attempts on the most-used axis is
$60\%/43\%/38\%$ for ADAS/AFlow/KDLoop, excluding no-change probes.
ADAS also repeatedly re-discovers the same MapGPT score,
$\mathrm{SR}{=}0.4769$, across six iterations, making further proposals without
finding a higher score in those iterations. These patterns suggest that
embodied AAS needs memory not only over scalar scores, but also over the
structure of attempted edits.

The resulting search problem is deciding when to refine an existing edit
family and when to explore a different one. Continued refinement can uncover
useful variations, but repeated attempts in the same direction consume budget
that could test other mechanisms. KDLoop's axis tags and coverage records make
these choices explicit, and its recorded attempts are less concentrated on a
single axis. The traces show different allocations of search effort, rather
than establishing which allocation is optimal. Tracking attempted edits and
their outcomes can inform this decision, but the optimizer must still judge
whether a plateau warrants further refinement or a change of direction.

\vspace{-2mm}
\subsection{Episode-level Credit Assignment Only Partially Emerges}
\label{sec:exp-credit}
\vspace{-2mm}

Prior AAS systems usually optimize opaque LLM modules against scalar metrics such
as accuracy or success rate. Embodied AAS provides richer evidence, including
observations, actions, tool calls, planner outputs, self-reports, and simulator
traces that can explain why a score changed. Our harness exposes these logs,
and all search agents know they are available. However, episode-wise credit
assignment only partially emerges. In the recorded searches, optimizers often
use aggregate scores to guide subsequent edits without tracing the change in
performance to specific episode-level behavior.

The add-one ablations provide a retrospective comparison with the optimizers'
own attribution during search (\S\ref{sec:exp-ablation}). For MapGPT, one edit
in each of the AFlow and KDLoop candidates accounts for most of its gain,
while other selected edits contribute little or reduce performance. The
optimizers did not make this distinction during search. Their selection of an
improved candidate therefore did not yield an equally specific account of
which edits were responsible, limiting the evidence available to guide the
next proposal.

The robustness probes expose related gaps in evidence use. In all VoxPoser
runs, we inject a silent framework-level defect in which the logs of
dynamically dispatched language-model program (LMP) sub-calls inside the
VoxPoser planner~\cite{huang2023voxposer} are missing from the voluntary
self-report channel. This controlled fault preserves scalar SR optimization
while testing whether optimizers inspect episode-level evidence. Although all
agents can inspect episode logs, ADAS and AFlow do not surface the missing
traces and remain focused on scalar evaluation outcomes. KDLoop is the only
variant that reports the missing traces in these runs, but it does so late
rather than as a routine attribution step. In SmartWay, AFlow obtains its best
score by wiring a ground-truth distance-to-goal signal from the Habitat evaluator
into the executable graph. The scalar metric rewards this change, but
episode-level inspection is needed to recognize that the mechanism depends on
evaluator-side information rather than a deployable reasoning pathway.

These cases show that log access alone is insufficient. Coding-agent-based AAS
does not reliably use episode-wise evidence for credit assignment, even when the
evidence is available and discoverable. Embodied AAS therefore needs explicit
attribution mechanisms that check whether a proposed graph mechanism actually
executed, whether the information it uses is deployable, and whether a failure
comes from the editable graph or the underlying substrate.

\vspace{-2mm}
\section{Limitations and Future Work}
\label{sec:limitations}
\vspace{-2mm}
This study provides an initial characterization of embodied AAS.
\textbf{1) Executable scope.} AgentCanvas represents executors as typed
node-and-wire workflows, and our experiments search around existing embodied
agents. Future work should extend AAS to richer executable representations
and safer edit operators beyond workflow-shaped agents.
\textbf{2) Optimization signal.} Costly, noisy rollouts can obscure the effect
of a graph edit, so the best single-pass score may not reliably estimate
architecture quality. Search dynamics can also induce local basins, where
variants repeatedly modify one promising mechanism rather than exploring
qualitatively different edits. KDLoop takes initial steps through typed
intervention history and space-exhaustion reflection, but future systems need
more adaptive, substrate-portable strategies for selection, coverage tracking,
and stopping.
\textbf{3) Credit assignment.} Success rate alone is insufficient for embodied
credit assignment. Although our harness exposes episode-level logs, detailed
episode-wise analysis only partially emerges. Future embodied AAS systems should
make attribution a first-class part of the loop, checking whether proposed
mechanisms execute, whether information sources are deployable, and whether
failures arise from the editable graph or the underlying substrate.

AAS can also support self-evolution and recursive self-improvement (RSI).
Extending search to the optimizer itself is a natural next step beyond our
current executor search.

\clearpage

\bibliography{references}

\clearpage
\appendix

\section*{Appendix Contents}
\begin{itemize}
  \item \textbf{Appendix~A~--~AgentCanvas.} The typed-graph Executor substrate
        (\S\ref{appx:agentcanvas}).
  \item \textbf{Appendix~B~--~Per-Cell Search Trajectories.} Full search
        trajectories across the 12-cell grid (\S\ref{app:trajectories}).
  \item \textbf{Appendix~C~--~Coding-Agent Harness.} The method-agnostic
        Optimizer-side substrate (\S\ref{appx:harness}).
  \item \textbf{Appendix~D~--~Implementation Details.} Per-variant proposer and
        memory (\S\ref{appx:impl-details}).
  \item \textbf{Appendix~E~--~Experiments Setup.} Executors, eval tiers, models,
        and the rerun protocol (\S\ref{appx:exp-setup}).
  \item \textbf{Appendix~F~--~Problem Formulation.} The formal search space,
        method-seeded neighborhood, admission, and objective
        (\S\ref{appx:formulation}).
  \item \textbf{Appendix~G~--~Held-out Evaluation.} Held-out split
        construction, the transfer protocol, and the paired significance test
        (\S\ref{app:heldout}).
  \item \textbf{Appendix~H~--~Search Replication and Cost.} Additional independent searches
        and recorded resource usage
        (\S\ref{app:search-robustness}).
\end{itemize}

\section{AgentCanvas: A Typed-Graph Executor Substrate for Embodied AAS}
\label{appx:agentcanvas}

This appendix documents \textsc{AgentCanvas}, the \textbf{Executor substrate}
introduced in \S\ref{sec:substrates}: the runtime that represents each embodied
agent as an editable, runnable, instrumented program and so makes the space
$\mathcal{C}$ of candidate architectures (\S\ref{sec:problem}) something an
Optimizer can actually traverse. Its scope here is deliberately narrow. We
describe only the designs that an AAS Optimizer relies on, not the human-facing
visual editor, the node-authoring API, or the wider platform. The full
implementation is in the public repository.
The Optimizer-side substrate, the coding-agent harness that proposes and
applies edits, is described in \S\ref{sec:substrates}. This appendix is its
Executor-side counterpart, and \S\ref{appx:ac-loop} ties the two together.

\S\ref{appx:ac-gap} states the two interface gaps in current embodied agents
that motivate the substrate, together with the class of agents the substrate
represents. \S\ref{appx:ac-graph}--\S\ref{appx:ac-validate} describe the
\emph{inference} interface: the agent as an editable typed graph and the
static checks that gate an edit before a rollout. \S\ref{appx:ac-eval} describes
the \emph{evaluate} interface, how an Optimizer autonomously scores a
candidate at benchmark scale, with guaranteed termination.
\S\ref{appx:ac-logs} describes the episode-level evidence available for credit
assignment, and its honest limit. \S\ref{appx:ac-loop} summarizes the substrate
as an AAS interface: the little the Optimizer must do, and the scaffolding it is
spared.

\noindent\textbf{Released code.}\ \ The full implementation is available at
\url{https://github.com/jianzhou0420/AgentCanvas}.

\begin{figure}[ht]
\centering
\includegraphics[width=\linewidth]{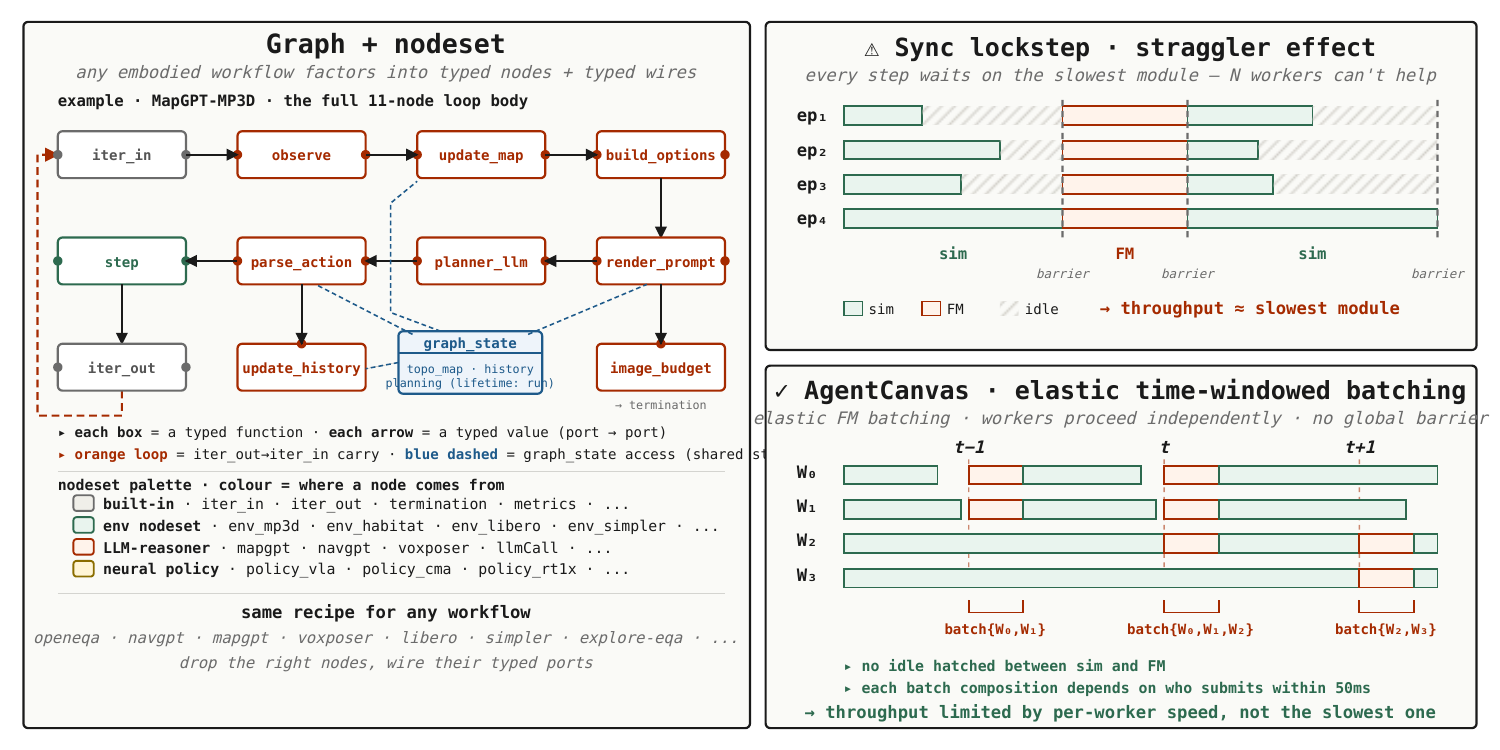}
\caption{\textbf{AgentCanvas as an embodied-AAS substrate.}
\emph{Left:} an agent factors into typed nodes, typed-port wires, and shared
state in a container reached by access grants (shown: the MapGPT-MP3D loop body;
node colour = source nodeset; the loop is an
\texttt{iter\_out}$\to$\texttt{iter\_in} carry, not a back-edge;
\S\ref{appx:ac-graph}--\S\ref{appx:ac-validate}).
\emph{Right:} the batch-eval optimisation (\S\ref{appx:ac-eval}). Naive
synchronous lock-step (top) stalls every worker on the slowest module, while
AgentCanvas (bottom) decouples workers and batches whichever FM calls arrive
within a short window (default 50\,ms), so throughput tracks per-worker speed.}
\label{fig:agentcanvas-substrate}
\end{figure}

\subsection{Two interface gaps in current embodied agents}
\label{appx:ac-gap}

Published embodied agents are typically shipped as bespoke, workflow-shaped
forward passes implemented in custom multi-file code: a navigation or
question-answering system hard-wires its own perception calls, prompt assembly,
memory updates, model placements, and action decoding into Python that is
specific to one method and one benchmark. For a \emph{human} researcher this is
legible enough. For an \emph{Optimizer} it presents two missing interfaces.

\textbf{(i) No unified inference interface.} Because each agent's structure lives
in idiosyncratic code, there is no common surface on which to express a
structural edit. ``Move the stop decision after the landmark check'', ``add a
voting node over three planner calls'', or ``re-point the observation encoder''
are uniform graph-level operations in principle, but in practice each requires
reading and rewriting method-specific code, and a malformed edit is only
discovered by running it. An Optimizer cannot apply structural edits uniformly,
nor cheaply reject an ill-formed one.

\textbf{(ii) No unified evaluate interface.} Each agent also carries its own
evaluation scaffolding, namely its own episode loop, its own simulator wiring, its
own metric bookkeeping. There is no standard call by which an external system
hands the agent a benchmark split and receives a success rate, and no standard
way to run that evaluation at the throughput a search loop needs. An Optimizer
cannot autonomously trigger scoring, and certainly cannot fan it out across
workers without re-engineering each agent's harness.

\begin{figure}[ht]
\centering
\includegraphics[width=\linewidth]{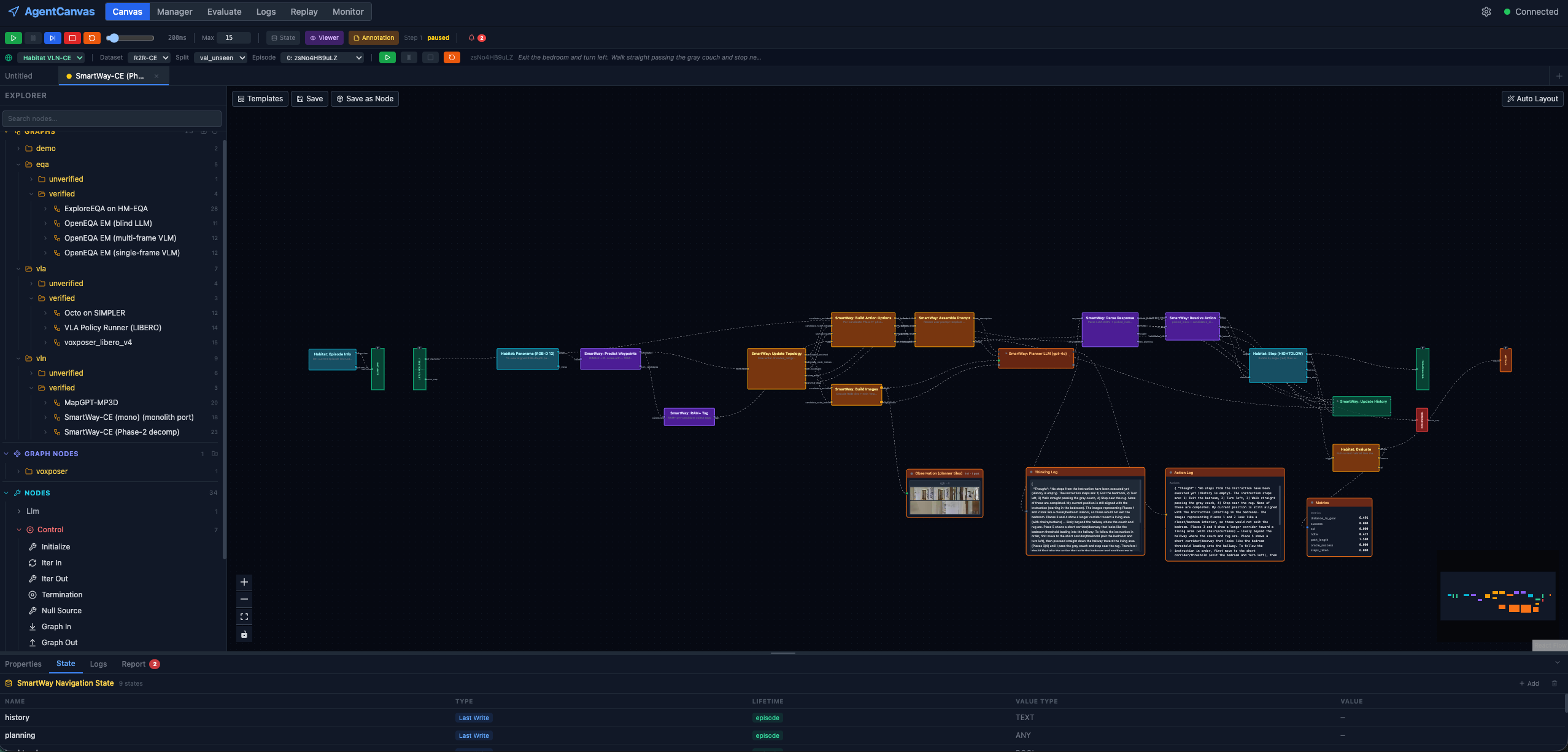}
\caption{\textbf{The AgentCanvas UI, for a human.} The typed graph an Optimizer
edits as JSON is, for a researcher, a node-and-wire canvas with an inspectable
state panel (shown: the SmartWay-CE graph and its run state). The visual editor
and the Optimizer's edit/evaluate interface are two views of one artifact.}
\label{fig:agentcanvas-canvas}
\end{figure}

\textsc{AgentCanvas} closes both gaps with a single substrate that is
simultaneously \emph{visual for a human} (a node-and-wire canvas,
Figure~\ref{fig:agentcanvas-canvas}) and
\emph{standard and batch-optimized for an Optimizer} (a typed-graph data model
with one structured-edit surface and one batch-eval surface). The substrate
represents an agent as a fixed forward graph over perception, memory, planning,
and action modules: this covers the published embodied methods we seed from
(\S\ref{sec:problem}) and is what makes method-seeded search tractable, but it
deliberately stops short of agents whose control flow, tool set, or memory are
\emph{constructed dynamically during deployment}, the boundary that becomes
Limitation~1 in \S\ref{sec:limitations}. The remainder of this appendix
describes the parts of that substrate that the Optimizer touches.

\subsection{The Executor as an editable typed graph}
\label{appx:ac-graph}

An Executor in \textsc{AgentCanvas} is a pure-data \texttt{GraphDefinition}: a
JSON document of typed \texttt{NodeDef}s, \texttt{EdgeDef}s, state containers,
and access grants that fully specifies one agent's topology and configuration
with no hidden runtime state (Figure~\ref{fig:agentcanvas-substrate}, left).\footnote{The data model lives in
\texttt{graph\_def.py}, and it round-trips through \texttt{GraphDefinition.from\_dict()}
/ \texttt{to\_dict()}, so an Optimizer can read, mutate, and write a candidate as
plain JSON.} Each \texttt{NodeDef} is a \emph{reference}, not an instance: its
\texttt{type} is a string key resolved at run time against the node registry
(\texttt{NODE\_HANDLERS} for built-ins, \texttt{\$\{nodeset\}\_\_\$\{node\}} for
nodeset-provided modules), with all per-instance knobs in a \texttt{config}
dictionary. Edges connect \emph{named} ports through
\texttt{sourceHandle}/\texttt{targetHandle}, and a region of the graph can be
wrapped into a reusable composite via \texttt{NodeDef.subgraph} with
\texttt{portIn}/\texttt{portOut} boundary nodes. The runtime engine
(\texttt{ReactiveExecutor}) expands composites with \texttt{flatten\_graph()}
before running, so nesting is an authoring convenience, not a runtime concept.

Because the whole agent is one referential artifact, the structural edits an AAS
proposal needs are exactly the legal mutations of this JSON: add, remove, or
retype a \texttt{NodeDef} (swap one module for another by changing its
\texttt{type}); rewire an \texttt{EdgeDef} between named ports; tune any node's
\texttt{config}; edit the state containers, access grants, \texttt{step\_budget},
or \texttt{terminationCondition}; or wrap/unwrap a subgraph. This is the surface
the Optimizer's implementer (\S\ref{sec:substrates}) writes against, and the
nodes that travel along it are themselves a uniform, swappable unit: every
node (a tool, a model call, an environment step, a whole sub-agent) is a
single class with declared \texttt{input\_ports} and \texttt{output\_ports} of
typed \texttt{PortDef}s and one \texttt{execute(inputs, ctx)} method, so module
substitution is type-shaped rather than code-shaped.

\emph{What this gives the Optimizer:} one artifact whose every structural choice
is an addressable JSON field, so a proposal is a structured patch rather than a
multi-file code rewrite, and ``one graph = one agent'' makes a candidate
portable, diffable, and loggable as a unit.

\subsection{Typed ports and pre-rollout validation}
\label{appx:ac-validate}

Ports carry a fixed wire-type catalog: perception payloads (images, depth
maps), control payloads (actions, poses), scalar and text payloads (text,
booleans, metric bundles), structured observation and step-result records, an
untyped escape hatch, and a \texttt{LIST[T]} modifier. Types are not decorative: they gate execution and, more importantly
for search, they gate \emph{admission}. Before any rollout,
\texttt{validate\_graph\_connectivity()} runs at graph load and rejects
(with an HTTP 400, never a silent never-fires) any candidate that wires
incompatible types, leaves a required (non-optional) input unconnected, mis-uses
a \texttt{LIST[T]} coercion, or violates the iteration-pivot wiring rules
(the \texttt{Initialize}/\texttt{iterIn}/\texttt{iterOut} pairing that expresses
a cyclic agent loop without graph back-edges). The legal edit and connection
space is itself
machine-readable: a node-schema endpoint exposes every node's ports and config
schema, so an Optimizer can enumerate what may be wired to what rather than
guessing.

\emph{What this gives the Optimizer:} a cheap, deterministic filter that
discards a structurally invalid proposal in milliseconds at load time, before it
spends a multi-episode GPU rollout, turning ``did my edit even type-check''
from a post-hoc rollout failure into a pre-flight static check.

\subsection{An autonomous, batch-optimized evaluate interface}
\label{appx:ac-eval}

The batch stack is shaped by an asymmetry between the two node kinds a run
contains. A foundation-model node (the LLM/VLM backbone) is \emph{stateless}
(inputs in, outputs out) and natively batchable, so its efficient
deployment is a single weights-loaded replica serving many callers at once. A
simulator node is \emph{stateful}: each parallel episode needs its own
simulator, so $K$-way evaluation spawns $K$ replicated simulator subprocesses
whose model calls all route to the one shared backbone. The naive way to batch
those calls (a fixed, lock-stepped batch that waits for all $K$ workers to
submit before firing one forward pass) reintroduces a straggler
(short-board) effect: every worker stalls on the slowest simulator step, so one
slow or long episode throttles the whole batch
(Figure~\ref{fig:agentcanvas-substrate}, right).

AgentCanvas avoids this with \emph{decoupled workers and an opportunistic,
time-windowed batch}. The $K$ workers are never lock-stepped: each drives its
own episode stream asynchronously (\texttt{EnvWorkerPool}), so a fast worker
advances to its next step or episode without waiting for the others. Inside the
shared backbone subprocess, a \texttt{BatchedInferenceServer} keeps one queue
per \texttt{(function, config)} key. Each caller submits a single sample and
awaits a future, and the queue flushes \emph{whatever has arrived} after a short
restart-on-submit debounce (\texttt{flush\_timeout\_ms}, default 50\,ms) rather
than after a fixed count. The batch size thus flexes to whoever is ready in the
window (a slow worker is simply absent from the current batch and joins a
later one instead of holding it open), and a single in-flight lock bounds peak
GPU memory to one batch while the next batch keeps accumulating. The model's
stateless contract is what makes this safe: any set of concurrent callers can be
stacked into one pass, with recurrent state, where present, carried explicitly
on the wire (\texttt{hidden\_in}/\texttt{hidden\_out}) rather than held in the
server.

A run is launched with one \texttt{POST /api/eval/v2/start} (graph, split,
episode count, \texttt{worker\_count}). A VRAM-admission scheduler queues
concurrent search sessions rather than letting them out-of-memory each other,
and the single-worker path is bit-identical to the parallel one, so throughput
scales without changing the measured score. The run returns aggregate success
rate (and SPL / nDTW where the benchmark defines them) in \texttt{summary.json},
plus a frozen \texttt{graph.json} and self-contained per-episode records.

\emph{What this gives the Optimizer:} a one-call, headless way to turn an edited
graph into a benchmark score that scales across workers without per-worker
weight reloads and without a straggler tax. This is what turns large-scale
embodied search from possible into routine.

\subsection{Episode-level evidence, and its honest limit}
\label{appx:ac-logs}

Every rollout is logged so that a score change can be \emph{read off} rather
than reproduced, and the log has two kinds of entries. The first is
\textbf{automatic}: for each node firing the framework records the node's inputs
and outputs, its timing and any error, and (for a model call) the model
name and token cost, with no effort from the node author. The second is
\textbf{optional}: a node may also record its own internals, the prompt it
assembled, the raw model reply, a planner decision, by calling
\texttt{\_self\_log}. Each episode is written to its own self-contained
\texttt{log.jsonl}, so an Optimizer can open episode~47 and see exactly what
fired, in order, without re-running or untangling it from other episodes.

The catch is that the second kind is opt-in. If a node does its real work
\emph{inside} itself (dispatching a tool call or a nested language-model
program) and does not \texttt{\_self\_log} it, that inner step never appears:
the log still shows the node's outer inputs and outputs, but not what happened
between them. So the evidence reliably pins a score change to a \emph{node}, but
cannot always see \emph{inside} one.

\emph{What this gives the Optimizer:} per-node, per-episode evidence that
localizes a score change to the firing that moved, with one stated blind
spot, an unlogged internal dispatch, where that evidence runs out.

\subsection{Summary: what the substrate asks of the Optimizer}
\label{appx:ac-loop}

Taken together, the preceding designs reduce embodied architecture search to a
loop with a tiny surface. To search, the Optimizer needs only three things from
the substrate: a typed JSON graph it can edit
(\S\ref{appx:ac-graph}--\S\ref{appx:ac-validate}), a single call that scores any
edited graph over a benchmark split (\S\ref{appx:ac-eval}), and the per-episode
logs that call returns (\S\ref{appx:ac-logs}). Edit the graph, issue one
evaluate call, read the result. That is the whole contract, and it is what
makes AAS over real embodied agents possible at all.

What the Optimizer therefore does \emph{not} have to do is the part that would
otherwise dominate. It does not read or maintain each agent's bespoke,
multi-file implementation. It does not write glue code to drive a simulator. And
it does not build an evaluation harness per method. This matters because the
Optimizer is itself a context-bound coding agent: when the agent under search is
ordinary code with its own runner, every iteration burns scarce context on
\emph{how to execute and score} a candidate. Collapsing the agent to a data
artifact behind a standard evaluator lets the Optimizer spend its context on
\emph{what to change} (the search itself) instead of on scaffolding.

And because that evaluator is the batched, straggler-avoiding stack of
\S\ref{appx:ac-eval}, each candidate is scored at low wall-clock, so the loop can
afford to evaluate many of them. A minimal edit surface, an autonomous and fast
evaluator, and readable per-episode evidence are jointly what turn embodied AAS
from a per-method engineering effort into a search an Optimizer can simply run.

\section{Full Per-Cell Search Trajectories}
\label{app:trajectories}

\begin{figure}[t]
\centering
\includegraphics[width=\linewidth]{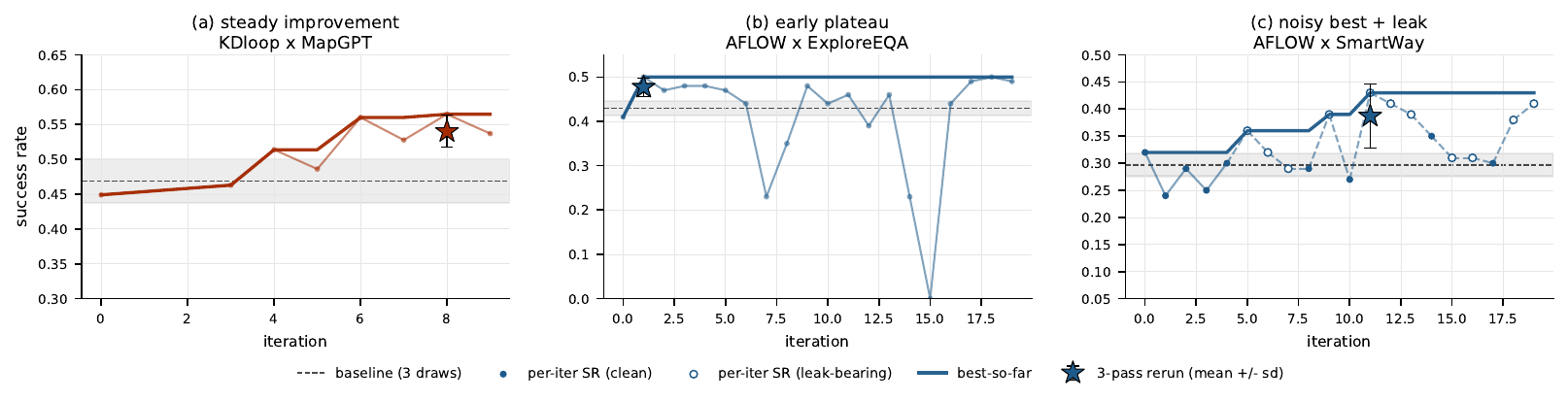}
\caption{Representative search trajectories. Each panel shows single-pass SR,
best-so-far SR, the shared baseline, and the rerun-confirmed best iteration.
The three cells illustrate improvement, plateau, and noisy best-selection; the
complete grid is Fig.~\ref{fig:trajectory-grid-full}.}
\label{fig:trajectories}
\end{figure}

\begin{figure}[t]
\centering
\includegraphics[width=\linewidth]{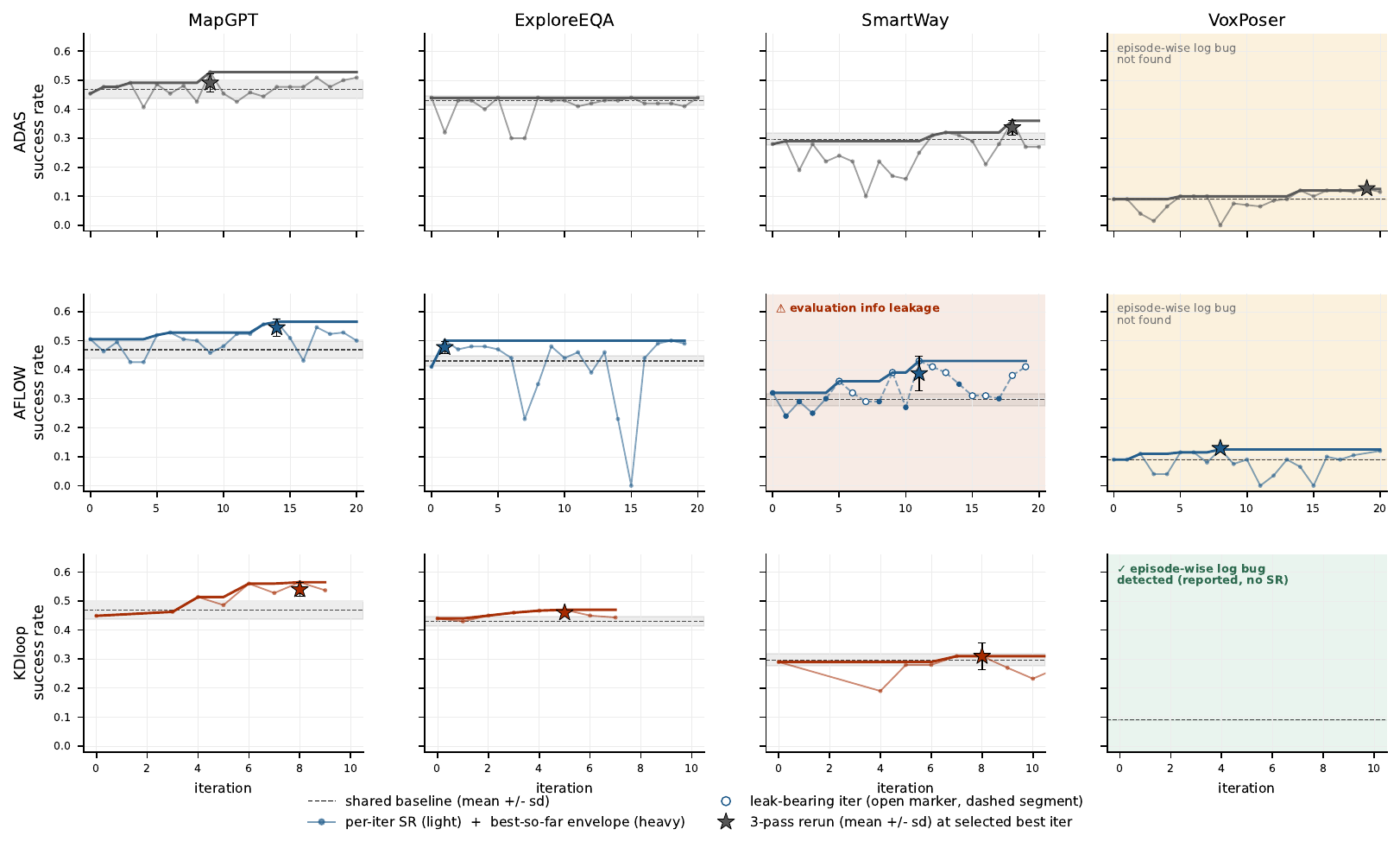}
\caption{Full $3\times4$ per-cell search trajectories (rows: optimizers,
columns: executors), same conventions as Fig.~\ref{fig:trajectories}:
per-iter single-pass SR (light line + markers), best-so-far envelope (heavy
line), shared baseline mean $\pm$\,sd (dashed line + band), and three-pass
rerun mean $\pm$\,sd at the selected best iter ($\star$). The success-rate
axis is shared across all panels. The iteration axis is $0$--$20$ for the
ADAS / AFlow rows and $0$--$10$ for KDLoop. In-cell shading marks the
substrate-level findings: KDLoop detecting the episode-wise logging bug
(green), ADAS / AFlow leaving it unfound (amber), and AFlow's SmartWay gain
riding on evaluation-information leakage (red). AFlow $\times$ SmartWay's leak
iters use dashed segments and open markers (\S\ref{sec:exp-credit}).}
\label{fig:trajectory-grid-full}
\end{figure}

ADAS and AFlow were designed as single-shot, non-agentic optimizers: one
meta-prompt emits an entire agent in a single shot, scored by cheap
exact-match on a text benchmark. That formulation does not run natively
inside a coding-agent session, and its text-domain assumptions (no runtime,
no episodes, near-free evaluation) do not transfer to embodied executors. We
lift all three optimizers onto one shared coding-agent harness
(\S\ref{sec:substrates}) with embodied adaptations, and treat them not as
competitors for a single ranking but as three \emph{search paradigms} for
embodied AAS (\S\ref{sec:variants}): 3-call Reflexion hill-climb (ADAS),
softmax-mix parent sampling (AFlow), and knowledge-distillation search
(KDLoop). The grid below is a \emph{qualitative} comparison. It shows all
three paradigms yield working search on embodied executors, each with a
characteristic strength and failure mode, and we do not expect KDLoop to dominate
search efficiency, effectiveness, and robustness at once. ADAS and AFlow
concentrate within one edit basin, which buys precise within-mechanism tuning,
e.g.\ AFlow on ExploreEQA spends eight of its nineteen search iters
refining a single answer-aggregator (relevancy-threshold $\to$ rel-weighted
sum $\to$ majority-vote $\to$ rel-gated mean-pool $\to$ Top-K confident-step),
plateauing near $0.50$. KDLoop instead searches several intervention axes in
parallel and designs its own targeted sub-experiments (e.g.\ failure-mode
episode subsets that accumulate reusable knowledge), giving more efficient
coverage. Its knowledge-driven search also avoids the evaluation-information
leakage AFlow falls into on SmartWay and surfaces the framework logging bug
on VoxPoser (\S\ref{sec:exp-credit}).
All trajectory scores here concern search episodes. Held-out performance and
additional independent searches are reported in Appendices~\ref{app:heldout}
and~\ref{app:search-robustness}.

Figure~\ref{fig:trajectory-grid-full} is the full $3\times4$ grid behind the
three representative cells of Fig.~\ref{fig:trajectories}. Plot conventions,
axes, and the in-cell shading are defined in the caption.

\paragraph{Special cells.} AFlow $\times$ SmartWay: $11$ leak-bearing iters
wire the evaluator's ground-truth distance-to-goal into the planner
(\S\ref{sec:exp-credit}), and the headline-selected iter\_11 is leak-bearing.
ADAS $\times$ ExploreEQA: no iter beats iter\_0 in $20$ iters, so no rerun
star. KDLoop $\times$ VoxPoser: exits with a substrate bug report, no SR.

\paragraph{Across the grid.} \emph{Steady growth} on MapGPT for all three:
AFlow peaks at iter\_14 ($0.545{\pm}0.031$ rerun), KDLoop at iter\_8
($0.540{\pm}0.023$), ADAS at iter\_9 ($0.491{\pm}0.032$), all above the
$0.469$ baseline. \emph{SmartWay} separates selection from verification: ADAS
$0.360{\to}0.337{\pm}0.025$ and AFlow $0.430{\to}0.387{\pm}0.059$ both shrink
on rerun, and AFlow's higher figure is leak-bearing (\S\ref{sec:exp-credit}),
whereas KDLoop holds a leak-free $0.310{\pm}0.046$. \emph{VoxPoser}: ADAS and
AFlow post marginal substrate-adjacent gains ($+0.037$ / $+0.039$), while
KDLoop diagnoses the episode-wise logging bug underlying the cell (reported,
no SR).

\section{Coding-Agent Harness as AAS Substrate}
\label{appx:harness}

This appendix documents the \textbf{method-agnostic coding-agent
harness} that hosts the AAS family. It is the substrate an AAS algorithm
runs on, not any algorithm itself. It is method-agnostic by
construction: swapping the algorithm replaces a single skill
(\texttt{proposer}) and reuses the rest. Concrete variants (ADAS,
AFlow, \ldots) and their variant-specific choices are out of scope.
Our \texttt{ADAS} port is in Appendix~\ref{appx:impl-details}.

\subsection{Pipeline overview}
\label{appx:overview}

One iteration factors into an orchestrator (\texttt{loop}) and four
workers: \texttt{understand} (a read-only context loader),
\texttt{proposer} (the one AAS-specific skill), \texttt{implementer},
and \texttt{evaluator}. The latter two delegate evaluation to a shared
runner (\texttt{/experiment:run}). Skills are Markdown prompts run by a
single coding-agent conversation per iteration. The decomposition
exploits a capability shift: where original ADAS \citep{hu2025automated}
needed bespoke implement-and-debug machinery (a typed edit language, a
deterministic replayer, repair loops) because models could not edit a
multi-file codebase from a spec, recent coding agents
\citep{anthropic2024claudecode, yang2024swe} can. We collapse that
whole branch into \texttt{implementer} (\S\ref{appx:impl}). The proposer
emits a prose change spec and a sub-agent edits the code natively.

\paragraph{Files contract.}
All variants share an on-disk contract
(\texttt{\_common/files-contract.md}). Runs live under a method-scoped
root \texttt{outputs/design\_runs/\{method\}/\{graph\}/v\{N\}/}, one dir
per iteration at \texttt{iteration/iter\_n/}. The one mandatory
per-iteration artifact is an \texttt{active\_workspace/} overlay
(\S\ref{appx:rundir}). Other artifacts are typed by a fixed taxonomy
and declared per variant in a manifest, so the substrate fixes
placement/mutability/lifecycle by type while each variant names only its
own files.

\subsection{The \texttt{loop} skill: per-iter orchestration}
\label{appx:loop}

\texttt{loop} owns one per-iteration conversation and drives
\texttt{proposer} $\to$ \texttt{implementer} $\to$ \texttt{evaluator},
with each worker writing into a transient staging directory. On
evaluator success it runs the atomic-commit transaction
(\S\ref{appx:rundir}). On any worker failure it discards staging,
increments \texttt{consecutive\_skips}, and treats the iteration as
never having happened. It terminates on an iteration cap, a user
\texttt{STOP} sentinel, or too many consecutive skips. Where the
algorithm needs independent samples or fresh tool contexts, a worker
issues \emph{sub-agent spawns}, self-contained sub-conversations
with full tool access that return a JSON block. $N$ spawns are $N$
independent samples, which a single thinking-pass (one shared
autoregressive trace) cannot provide. The substrate supplies the spawn
primitive, and how many to issue and how to combine them is the variant's
choice (Appendix~\ref{appx:impl-details}).

\subsection{Implementer: native editing and the smoke gate}
\label{appx:impl}

The proposer hands over a change spec, a prose \texttt{intent} plus a
\texttt{targets} file list, not a typed op list. A deterministic helper
(\texttt{\_common/lib/overlay.py}) seeds each target into the overlay
and enforces the edit whitelist (a hard wall around framework and
vendored code, and a soft warn for edits outside the iteration's graph and
nodesets), but never edits. A tool-augmented sub-agent then edits the
seeded files natively. The implementer validates the result (JSON still
parses, Python still compiles) and pins every model-call node to one
enforced inference profile (\texttt{pin\_llm\_profile.py}) so the search
cannot vary the underlying model. It then runs a small smoke evaluation
(default $5$ episodes, profile \texttt{smoke\_\{graph\}}) whose gate is
\textbf{runtime correctness only}: pass iff the run exits cleanly, every
episode completes, takes $\geq 1$ step, and returns a valid numeric
metric. The metric \emph{values} are not consulted, so a clean
zero-scoring run is data, not failure. On any failure the overlay
resets to the parent and a \emph{fresh} sub-agent retries (default
\texttt{retry\_max}${=}3$), seeing only the \texttt{intent} plus failure
traces as read-only context, with no inherited message history. On
exhaustion the implementer reverts and signals a skip, and frozen
\texttt{workspace/} is never touched.

\subsection{Evaluator and the fitness signal}
\label{appx:eval}

The evaluator runs once per successful implementer on a separate profile
(\texttt{perf\_\{graph\}}, typically $100$ episodes) and writes a
\textbf{neutral} metrics record (run id, episode count, per-episode
accuracy list, primary-metric mean, secondary metrics). It is
\textbf{method-free}: how those numbers become the optimizer's scalar
fitness (a mean, a $t$-interval, or our ADAS port's bootstrap
confidence interval) is a variant choice computed in the variant's
\texttt{loop} during commit (Appendix~\ref{appx:impl-details}), which is what keeps the
evaluator reusable across variants. The cheap-smoke / expensive-perf
split is cost-driven: a $100$-episode run costs $1$--$2$ hours, so
retries are confined to the smoke tier, and a low-but-valid perf result is
recorded, not retried.

\subsection{Run-directory contract and atomic commit}
\label{appx:rundir}

All state is filesystem-based. Frozen \texttt{workspace/} is never
written by any skill, and an iteration's edits land only in its
\texttt{active\_workspace/} (bootstrapped from the parent, then patched),
which the backend overlays over the frozen baseline at evaluation time
(overlay wins on conflict). This overlay is the single point coupling
the file contract to runtime behaviour. It lets each iteration, and
concurrent runs over one baseline, evaluate independent edits without a
per-iteration checkout. Workers write to staging
(\texttt{v\{N\}/.staging/iter\_n/}). On success the orchestrator moves
it into \texttt{iteration/} and \emph{then} appends one entry to the
variant's working memory (for ADAS, one archive line enriched with the
fitness statistic). The move-then-append order makes a failed iteration
a clean no-op: a half-written iteration never enters the working memory.

\section{Variant Implementation Details}
\label{appx:impl-details}

The three variants share the substrate of Appendix~\ref{appx:harness} and differ
only in the proposer skill and the memory that persists across iterations
(\S\ref{sec:variants}). Where \S\ref{sec:variants} states \emph{what} each port
preserves from upstream, this appendix gives \emph{how} our port realizes it. The
released code for all three lives under \texttt{.claude/commands/architect/}, and
each algorithm's caption names its specific files, so the prose below stays at the
level of mechanism. All three algorithms share one skeleton: a one-line setup
then a flat per-iteration loop. In the loop body, \Call{Implement}{} (native
editing $+$ smoke gate) and \Call{Evaluate}{} (neutral metrics) are the shared
substrate, and everything else is variant-specific.

\subsection{ADAS}
\label{appx:impl-adas}

We realize each of upstream's three Reflexion calls
\citep{hu2025automated} as an \emph{independent} tool-augmented sub-agent spawn,
recovering per-call sampling diversity while giving every sample full
read/grep/shell access. The three spawns share one message history rendered as
text, and the deliverable is a prose change spec (an \emph{intent} plus a list of
target files). We pre-seed the archive with the baseline plus seven reference
patterns as text-only, null-fitness entries, a design palette to adapt, not
benchmarked alternatives. Two upstream mechanisms are deliberately dropped: the
stateless-API sampling knobs (no analogue once a call is a conversation), and the
smoke gate's ``mean\,${<}0.01\Rightarrow$ debug'' trigger (a clean zero-scoring
embodied run is data, not a bug, so the gate checks runtime correctness only).
Fitness is a bootstrap CI computed in the variant loop at commit, keeping the
shared evaluator method-free. The Reflexion-only reflection field is stripped
before the archive append.

\begin{algorithm}[ht]
\caption{ADAS port. Files: \texttt{adas-subagent/\{loop,proposer\}.md},
\texttt{lib/helpers.py}; substrate \texttt{\_common/\{implementer,evaluator\}.md}.}
\label{alg:adas}
\footnotesize
\begin{algorithmic}[1]
\State \Call{PreSeed}{}: baseline $\textit{iter}_0$ $+$ $7$ reference patterns (\textit{fitness}${=}$null) \Comment{loop \S3}
\For{each iteration $n$ \textbf{until} cap / STOP / skips} \Comment{loop \S4}
  \State \textit{analysis} $\gets \Call{AnalyzeHead}{\textit{evaluatedArchive}[-1]}$ \Comment{proposer \S2, skip reference entries}
  \State \textit{base} $\gets \Call{BuildPrompt}{\textit{archive},\, \textit{analysis}}$ \Comment{\S3--4: full archive injected}
  \State $s_0 \gets \Call{Spawn}{\textit{base}}$ \Comment{\S6 call \#1: propose}
  \State $s_1 \gets \Call{Spawn}{\textit{base},\, s_0,\, \textsc{Reflexion}_1}$ \Comment{call \#2 (shared history)}
  \State $s_2 \gets \Call{Spawn}{\textit{base},\, s_0,\, s_1,\, \textsc{Reflexion}_2}$ \Comment{call \#3 $\to$ patch}
  \State \textbf{if} any spawn fails \textbf{then} rollback; \textbf{continue}
  \State \textbf{if} $\neg\Call{Implement}{s_2.\textit{patch}}$ \textbf{then} rollback; \textbf{continue} \Comment{smoke, $\le 3$ retries}
  \State \textit{acc} $\gets \Call{Evaluate}{\texttt{perf}}$ \Comment{neutral metrics}
  \State \textit{fit} $\gets \Call{BootstrapCI}{\textit{acc}}$ \Comment{in loop, not evaluator}
  \State \Call{Commit}{}: append \textit{archive} (strip \textit{reflection})
\EndFor
\end{algorithmic}
\end{algorithm}

\subsection{AFlow}
\label{appx:impl-aflow}

Each iteration first selects a parent from the archive by score-softmax mixing, so
any prior iteration can seed the next edit. We follow upstream's \emph{code}
defaults ($\alpha{=}0.2,\lambda{=}0.3,K{=}4$), not the paper's Eq.~3
($\alpha{=}0.4,\lambda{=}0.2$) \citep{zhang2025aflow}. The parent's prior accepted
and rejected modifications are injected as ``avoid-repeating'' experience. A
duplicate (under whitespace-normalized matching) resamples the parent, and the
resample loop (unbounded upstream) is capped at five. The search space is
free-form structural edits over the typed Executor graph (no operator library),
and the selected parent's overlay is checked out into staging before
\Call{Implement}{}.
Two further deviations, both on the evaluation side. (i)~Cost rules out the
five-pass validation average, so each iteration is scored on a \emph{single}
performance pass. Because that score feeds both the softmax ranking and the
convergence test, the unmodeled run-to-run variance propagates into the search
itself, not just the final report, and we calibrate it with post-loop
verification reruns on the top iterations. (ii)~Under single-pass scoring every
per-round variance is zero, so upstream's top-$3$ convergence predicate collapses
to exact equality regardless of $z$. Memory is the ADAS archive
plus a parent id, the verbatim modification string, and a bare numeric score (the
median of bootstrap means) for the softmax.

\begin{algorithm}[ht]
\caption{AFlow port. Files: \texttt{aflow/\{loop,proposer\}.md},
\texttt{lib/aflow\_helpers.py}; substrate as ADAS.}
\label{alg:aflow}
\footnotesize
\begin{algorithmic}[1]
\State \Call{PreSeed}{}: baseline $\textit{iter}_0$ only \Comment{no reference palette}
\For{each iteration $n$ \textbf{until} cap / STOP / skips / converged}
  \State \textit{experience} $\gets \Call{ExperienceMap}{\textit{archive}}$ \Comment{parent $\mapsto$ succ/fail mods}
  \For{$\textit{attempt} = 1 \ldots 5$} \Comment{anti-replay}
    \State \textit{top} $\gets \Call{TopRounds}{\textit{archive},\, K{=}4}$ \Comment{$\textit{iter}_0$ forced in}
    \State $p \gets \Call{SelectRound}{\textit{top},\, \alpha{=}.2,\, \lambda{=}.3}$ \Comment{softmax-mix}
    \State $r \gets \Call{Spawn}{\Call{OptimizePrompt}{p,\, \textit{experience}[p]}}$ \Comment{single call}
    \State \textbf{if} $\neg\Call{Parsed}{r}$ \textbf{then continue} \Comment{regex fallback}
    \State \textbf{if} $\Call{CheckMod}{r,\, \textit{experience}[p]}$ \textbf{then continue} \Comment{duplicate}
    \State \textbf{break} \Comment{novel modification}
  \EndFor
  \State \textbf{if} none accepted \textbf{then} rollback; \textbf{continue}
  \State write \texttt{proposal.md}$(p,\, r.\textit{mod})$; \ \Call{Checkout}{$p$} \Comment{loop \S4b}
  \State \textbf{if} $\neg\Call{Implement}{r}$ \textbf{then} rollback; \textbf{continue}
  \State \textit{acc} $\gets \Call{Evaluate}{\texttt{perf}}$ \Comment{every iteration}
  \State \textit{score} $\gets \Call{BootstrapMedian}{\textit{acc}}$ \Comment{feeds \Call{SelectRound}{}}
  \State \Call{Commit}{}$(p,\, \textit{mod},\, \textit{score})$; \ \Call{CheckConvergence}{} \Comment{stop if converged}
\EndFor
\State \Call{VerifyReruns}{top-1, top-2, baseline} \Comment{loop \S8}
\end{algorithmic}
\end{algorithm}

\subsection{KDLoop}
\label{appx:impl-kdloop}

KDLoop is the variant we design for the embodied setting, where each iteration
yields more evidence than a scalar score. It replaces the single propose step
with a four-phase cycle plus a triggered meta-phase. \textsc{Think} (one
tool-augmented spawn) reads the typed memory and emits up to three experiment
specs, each tagged with an intervention axis. An empty \textsc{Think} result
does not itself stop the run but escalates to \textsc{Reflect}. \textsc{Critic} spawns once
per patched spec to predict whether it would re-introduce a previously refuted
pathology, with one rebuttal round back through \textsc{Think}.
\textsc{Experiment} (inline, no separate implementer skill) applies each
surviving spec in its own overlay, then submits all \emph{(spec, pass)} pairs as
\emph{one} parallel wave whose per-submission worker counts are set by a makespan
minimizer under the performance-tier worker cap. \textsc{Distill} (one spawn over
all specs) writes lessons back to memory. \textsc{Reflect} is a meta-phase that
fires only on a three-iteration heartbeat, on the last three iterations sharing
one axis, or on a \textsc{Think} escalation, and it audits search-space coverage.
Memory is eleven typed files rather than a flat archive. The load-bearing ones are
pure facts, an open-hypothesis queue, an append-only log of closed-case lessons
(confirmed \emph{and} refuted), a coverage map, an eval-profile registry, and
self-authored utilities. These files are mutated by the explicit eager writes
shown in Alg.~\ref{alg:kdloop}: \textsc{Think} appends new conjectures to the
hypothesis queue, and for every hypothesis that its experiment resolves,
\textsc{Distill} appends a lesson to the experience log and line-deletes the entry
from the queue, so a refuted patch is committed as a lesson rather than
discarded. Termination is goal-driven or a \textsc{Reflect} ``space-exhausted''
verdict, subject to iteration and consecutive-skip limits.

\begin{algorithm}[ht]
\caption{KDLoop. Files: \texttt{myloop/loop.md} and phase skills
\texttt{\{proposer,critic,distill,reflect\}.md}, \texttt{lib/multi\_spec\_eval.py}.}
\label{alg:kdloop}
\footnotesize
\begin{algorithmic}[1]
\Statex \textit{Typed memory} $M$: \texttt{knowledge.md}, \texttt{hypotheses.jsonl}, \texttt{experience.jsonl}, \texttt{search\_space.md}, \ldots\ \ ($\mathrel{+}{=}$ append, $\mathrel{-}{=}$ line-delete)
\For{each iteration $n$ \textbf{until} goal met / space exhausted / cap / skips} \Comment{\S3a: poll \texttt{goal.md}}
  \If{\Call{ReflectTrigger}{}} \Comment{heartbeat-$3$ / axis-$3$}
    \State $v \gets \Call{Reflect}{M}$; \ \texttt{search\_space.md}\,$\mathrel{+}{=}$\,coverage \Comment{eager}
    \State \textbf{if} $v{=}$exhausted \textbf{then stop} (\textsc{saturated})
  \EndIf
  \State \textit{specs} $\gets \Call{Think}{M}$ \Comment{$K{\le}3$ axes}
  \State \quad \texttt{hypotheses.jsonl}\,$\mathrel{+}{=}$\,new conjectures \Comment{THINK eager}
  \State \quad \texttt{knowledge.md},\ \texttt{experiment\_design.yaml},\ \texttt{tools/}\,$\mathrel{+}{=}\ldots$
  \If{\textit{specs}${=}\emptyset$}
    \State \textbf{if} \textsc{Reflect} already ran this iteration \textbf{then} skip iteration
    \State $v \gets \Call{Reflect}{M}$
    \State \textbf{if} $v{=}$exhausted \textbf{then stop}
    \State \textbf{if} $v{=}$frontier-open \textbf{then} $\textit{specs} \gets \Call{Think}{M}$ \Comment{one retry}
    \State \textbf{if} \textit{specs}${=}\emptyset$ \textbf{then} skip iteration
  \EndIf
  \For{\textit{spec} $\in$ \textit{specs} \textbf{with} patch} \Comment{\S3b.5 CRITIC}
    \State $c \gets \Call{Critic}{\textit{spec}}$
    \State \textbf{if} $c \in \{\textsc{revise},\textsc{block}\}$ \textbf{then} \textit{spec} $\gets \Call{Think}{\textit{spec},c}$; recheck \Comment{1 rebuttal}
  \EndFor
  \State \textit{ready} $\gets$ surviving patched specs $\cup$ no-patch specs
  \For{\textit{spec} $\in$ \textit{ready} \textbf{with} patch} \Comment{\S3c: apply, own overlay}
    \State seed overlay; \Call{Implement}{\textit{spec}} \Comment{no stacking}
  \EndFor
  \State \textit{meta} $\gets \Call{MultiSpecEval}{\textit{ready}}$ \Comment{allocates workers before one wave}
  \State \textit{verdicts} $\gets \Call{Distill}{\textit{ready},\, \textit{meta}}$ \Comment{one spawn}
  \For{each hypothesis $h$ resolved this iter} \Comment{distill.md eager}
    \State \texttt{experience.jsonl}\,$\mathrel{+}{=}$\,lesson$(h)$ \Comment{confirmed / refuted / inconcl.}
    \State \texttt{hypotheses.jsonl}\,$\mathrel{-}{=}$\,$h$ \Comment{resolved $\to$ off queue}
  \EndFor
  \State \texttt{hypotheses.jsonl}\,$\mathrel{+}{=}$\,new; \ \texttt{knowledge.md}\,$\mathrel{+}{=}$\,facts \Comment{DISTILL eager}
  \State \Call{Commit}{\texttt{record.json}} \Comment{\S5}
\EndFor
\end{algorithmic}
\end{algorithm}

\section{Experiments Setup Detail}
\label{appx:exp-setup}

This appendix expands the setup of \S\ref{sec:experiments}. Every cell of the
$3\times4$ matrix runs on the same harness (Appendix~\ref{appx:harness}) and the
same per-executor evaluation profile, and variants differ only in the proposer and
memory (Appendix~\ref{appx:impl-details}).

The main matrix uses a single pinned AgentCanvas build, with executors
represented as forward node-and-wire graphs.

\subsection{Executors and benchmarks}
\label{appx:exp-executors}

Each cell seeds search from a published embodied-agent method and scores
candidates by task success rate (SR) on that method's benchmark
(Table~\ref{tab:exp-setup}). The performance tier is the paper-comparable eval
set and supplies every reported SR. The navigation cells additionally log SPL
and (n)DTW as secondary metrics, EQA logs step count, and manipulation logs step
count and cumulative reward.

\begin{table}[h]
\centering\small
\setlength{\tabcolsep}{5pt}
\begin{tabular}{@{}lllll@{}}
\toprule
Executor & Domain & Simulator & Split & Perf.\ episodes \\
\midrule
MapGPT     & VLN (discrete)    & Matterport3D     & \texttt{MapGPT72}    & $216$ (val\_unseen subset) \\
SmartWay   & VLN (continuous)  & Habitat (VLN-CE) & \texttt{rand100}     & $100$ (val-unseen subset) \\
ExploreEQA & Embodied QA       & HM3D             & \texttt{val}         & $100$ (of $500$ HM-EQA val) \\
VoxPoser   & VLA manipulation  & LIBERO           & VAS                  & $200$ ($4$ suites $\times 10 \times 5$) \\
\bottomrule
\end{tabular}
\caption{The four executors and their performance-tier evaluation sets. VAS is
the four LIBERO suites (\texttt{spatial / object / goal / 10}), ten tasks each,
five episodes per task. SR is \texttt{success} on each benchmark's own scorer.}
\label{tab:exp-setup}
\end{table}

\subsection{Evaluation tiers}
\label{appx:exp-tiers}

Each executor profile defines a cheap deterministic \emph{smoke} tier
($3$--$8$ episodes) and a paper-comparable \emph{performance} tier
(Table~\ref{tab:exp-setup}). Every candidate is smoke-gated before it is ever
measured. The smoke tier checks runtime correctness only (the run exits cleanly,
each episode takes $\geq 1$ step and returns a valid metric) and a candidate that
fails it is repaired or skipped, never scored (Appendix~\ref{appx:impl}). The two
ADAS-family ports then measure each surviving candidate on the \emph{full}
performance tier: AFlow on every iteration, because its score-softmax parent
sampling needs a comparable per-iteration score, and ADAS once per successful
candidate, for its archive and report. KDLoop instead \emph{designs} its own
experiment each iteration. Its Think phase composes a targeted subset, and a
$\geq 30$-episode failure-mode set is the default measurement tier. It escalates
to the full performance tier only once a custom run shows non-degenerate lift
($\geq 0.05$ SR over the prior best), to amortize the $1$--$2$\,h cost. Episode
budgets and worker counts are pinned per executor (e.g.\ MapGPT $15$ env steps at
$120$\,s each, SmartWay $20$ at $180$\,s, and VoxPoser $200$ waypoints).

\subsection{Models and iteration budget}
\label{appx:exp-models}

The searched agents use a single pinned backbone, \textsc{gpt-5-mini} at
temperature $1$, deterministically enforced on every model-call node
(\S\ref{appx:impl}). The VoxPoser composer is the exception, run on
\textsc{gpt-4o}. The search harness is fixed throughout: Claude Code provides the
coding-agent session and Claude Opus~4.7 with a 1M-token context is the
orchestrator. ADAS and AFlow run $20$ search iterations per cell and KDLoop runs $10$, not
counting the shared \texttt{iter\_0} baseline each evaluates first. Because each
KDLoop iteration designs and runs several experiments, its ten iterations are
roughly equivalent in compute to twenty ADAS/AFlow iterations, so the per-cell
budget is comparable across optimizers.
Actual resource usage is reported in Appendix~\ref{app:search-robustness}. The main matrix uses a single run per
(optimizer, executor) cell, with no multi-seed averaging, and variance is reported
within a run (next).

\subsection{Baseline and rerun protocol}
\label{appx:exp-rerun}

To separate search-time selection from rerun variance we report two reruns, both
on the performance tier. (i)~\emph{Shared baseline}: the three optimizers each
evaluate the same baseline graph as their \texttt{iter\_0}, giving three
independent SR draws per cell, and their mean and sample standard deviation are the
search-set baseline rows of Table~\ref{tab:headline}. (ii)~\emph{Best-iter rerun}: each
cell's selected best iteration is re-evaluated three times under the identical
performance config. We report the mean$\pm$sd and the gain $\Delta$ against the
shared-baseline mean.
For VoxPoser, episodes lost to simulator-server disconnects (at most ten of
$200$ per pass) are counted as failures.

\subsection{VoxPoser logging fault}
\label{appx:exp-voxposer-fault}

All VoxPoser runs share one controlled substrate-level fault: dynamically
dispatched LMP sub-calls are omitted from the agent's voluntary self-report
channel. The fault does not block scalar SR optimization, so it acts as a
diagnostic of whether an optimizer inspects episode-level evidence rather than
only the scalar outcome. In our runs this cleanly separates the three optimizers:
all share the same log access, yet ADAS and AFlow never surface the missing traces
and keep optimizing the scalar SR, while KDLoop is the only variant to inspect the
per-episode logs and report the fault (\S\ref{sec:exp-credit}).

\section{Problem Formulation for Embodied AAS}
\label{appx:formulation}

Automated agent design has produced almost as many problem formulations as
methods, because a formulation tends to fold its method's own commitments into the
statement of the problem, so what reads as the problem is really one method's
move-set. And because the field is text-agent dominant, those formulations are
shaped for prompt-and-tool pipelines, not for the closed perception--action loops
of embodied agents, so they do not transfer. We therefore give a formulation
that is both embodied and method-agnostic. We keep the template ADAS \citep{hu2025automated} adopts by
analogy to neural architecture search (a \emph{search space}, a \emph{search
algorithm}, and an \emph{evaluation function}) but draw the line the template
leaves implicit, between the problem and the method. The \emph{problem} is the
search space together with the objective over it: fixed, method-agnostic, and
shared by every variant (\S\ref{appx:f-substrate},~\S\ref{appx:f-eval}). The
\emph{method} is the search algorithm, the transition rule that produces the
next candidate, the only component that varies (\S\ref{sec:problem}). We
formalize it for the three variants we study: ADAS, AFlow, and our KDLoop
(\S\ref{appx:f-optimizer}).

\subsection{AgentCanvas as substrate: the search space}
\label{appx:f-substrate}

The substrate fixes what a candidate is and which candidates are well-formed. It
is what \emph{defines} the search space rather than shrinking a pre-existing one
(\S\ref{appx:ac-gap}). We give the candidate object and the operators that edit it,
then the space in two views (its \emph{open form}, the set of all expressible
candidates independent of any starting point, and its \emph{method-seeded} region,
what the search reaches from a published baseline) and close with what the space
can express.

\paragraph{Candidate.}
A candidate is a typed, config-bearing graph,
\[
  c=(V,\,\Sigma,\,E,\,\gamma),\qquad E=E_{\mathrm{flow}}\sqcup E_{\mathrm{mem}},
\]
with four constituents: the \emph{nodes} $V$, drawn from an open,
optimizer-extensible node library $\mathcal{N}$; the \emph{memory cells} $\Sigma$,
the candidate's declared shared state; the \emph{edges} $E$, of two kinds, namely
dataflow edges $E_{\mathrm{flow}}$ and memory edges $E_{\mathrm{mem}}$; and the
\emph{configuration} $\gamma$, a per-node parameter map. We expand each in turn.

\emph{Nodes.} A single graph holds finitely many nodes ($V\in\mathcal{N}^{*}$),
but $\mathcal{N}$ is not a fixed palette: the optimizer may author a new node type
and modify or delete an existing one, realized as patches that add or rewrite a
\texttt{BaseCanvasNode} subclass. Formally
$\mathcal{N}=\mathcal{N}_0\cup\mathcal{N}_\Delta(c)$, a seed library
$\mathcal{N}_0$ plus the node types defined within $c$'s own overlay, so the
library is endogenous to $c$ rather than an exogenous constant.

\emph{Memory cells.} Each cell $\sigma\in\Sigma$ holds a typed value with a
\emph{reducer} (accumulator, last-write, or counter) that combines concurrent
writes and a \emph{lifetime} that fixes when a reset signal clears it.

\emph{Edges.} Dataflow edges $E_{\mathrm{flow}}$ wire nodes to nodes and must be
type-compatible. Memory edges $E_{\mathrm{mem}}\subseteq V\times\Sigma$ connect a
node to a cell: $(v,\sigma)\in E_{\mathrm{mem}}$ lets node $v$ read and write cell
$\sigma$. A memory edge is of a different kind from a dataflow edge (it carries
no payload and does not trigger firing), and the two kinds are disjoint by
endpoint type, so a dataflow payload may not live in a cell.

\emph{Configuration.} The per-node map $\gamma$, with
$\gamma(v)\in\mathrm{Cfg}(t(v))$, gives each node its prompt text, sampling
parameters, and \texttt{persist} flags. ($\Sigma$ and
$E_{\mathrm{mem}}$ are the \emph{declared, shared} memory. A node's private
runtime scratch and an \texttt{iterIn} \texttt{persist} slot (carried dataflow,
held in $\gamma$) are not part of it.)

\emph{Everything is a node.} Two things that a formulation often leaves outside
the candidate are here ordinary nodes in $V$. The first is control flow: the loop
and its stopping rule are not an external harness but the nodes
\textsc{Initialize}, \textsc{iterIn}, \textsc{iterOut}, and \textsc{Termination}.
The second is the environment: the simulator the agent acts in is itself a node,
joined to the agent by ordinary dataflow edges. The agent reads observations
from it and sends actions back. A candidate is therefore one self-contained,
closed-loop graph holding the agent, its control flow, and its environment
together, with no hidden code running outside it. (The environment node supplies
the task, observations in and actions out, and is held fixed across a run. What
the search varies is the agent and how it is wired to that node.) A candidate's
\emph{coarse procedure}, how it plans and how it acts
each step, is thus just its topology, interior to $c$ and part of what is
searched, not a knob of the algorithm. This is the data model of
Appendix~\ref{appx:agentcanvas} (repro anchor \texttt{graph\_def.py}).

\paragraph{Operators.}
A candidate of the shape just defined comes with its move-set for free. The
constituents that can be edited are fixed (the node-type library
$\mathcal{N}_\Delta$, the node instances $V$, the memory cells $\Sigma$, the two
edge kinds $E_{\mathrm{flow}}$ and $E_{\mathrm{mem}}$, and the configuration
$\gamma$), so the high-level operators are simply one add / modify / remove
family per constituent, a finite alphabet $\mathcal{O}$:
\begin{center}\small
\begin{tabular}{@{}llll@{}}
\toprule
constituent & add & modify & remove\\
\midrule
node type $\mathcal{N}_\Delta$ & \textsc{define-type} & \textsc{modify-type} & \textsc{delete-type}\\
node instance $V$ & \textsc{insert-node} & \textsc{retype-node} & \textsc{remove-node}\\
memory cell $\Sigma$ & \textsc{add-cell} & \textsc{edit-cell} & \textsc{remove-cell}\\
dataflow edge $E_{\mathrm{flow}}$ & \textsc{connect} & \textsc{rewire} & \textsc{disconnect}\\
memory edge $E_{\mathrm{mem}}$ & \textsc{grant} & --- & \textsc{revoke}\\
config $\gamma$ & \multicolumn{3}{l}{\textsc{set-config} (fields: prompt / param / persist)}\\
\bottomrule
\end{tabular}
\end{center}
\textsc{define-type} extends the \emph{vocabulary} $\mathcal{N}_\Delta$, while
\textsc{insert-node} instantiates an existing type into the \emph{graph} $V$.
These are distinct operators, since authoring a new node and using it are separate moves.
\textsc{edit-cell} retypes a cell or changes its reducer or lifetime, and a memory
edge has no modify form because it confers read and write together.

This alphabet is \emph{complete} at the structural level. Any finer,
implementation-level edit (rewording a line of a prompt, flipping one
\texttt{persist} flag, swapping a tool inside a node) is a special case of one
of these operators (here, \textsc{set-config}). It changes a constituent's
contents without introducing a new \emph{kind} of move. The alphabet closes for a
structural reason: it is generated by the constituents of $c$ themselves, and an
edit touching nothing in $\{\mathcal{N}_\Delta,V,\Sigma,E,\gamma\}$ would not be an
edit to a candidate at all. The operators are afforded by the substrate's data
model, and the search policy (\S\ref{appx:f-optimizer}) is what realizes each as a
code diff and sequences them.

\paragraph{The open-form space.}
Let $\mathcal{W}(V,\Sigma)$ denote the well-typed edge sets over $V\cup\Sigma$
(type-compatible dataflow edges, plus memory edges to existing cells) and
$\mathrm{Valid}(\cdot)$ the static well-formedness check of
\S\ref{appx:ac-validate} (well-typed edges of both kinds, three-pivot coherence,
reachable graph in/out). The open-form search space is the set of all valid
candidates:
\begin{equation}
  \mathcal{C}
  =
  \Big\{\,
    c=(V,\Sigma,E,\gamma)
    \ \Big|\
    V\in\mathcal{N}^{*},\,
    E\in\mathcal{W}(V,\Sigma),\,
    \gamma\in\!\!\prod_{v\in V}\!\mathrm{Cfg}\big(t(v)\big),\,
    \mathrm{Valid}(c)
  \,\Big\}.
  \label{eq:space-declarative}
\end{equation}
It is fixed by membership alone (a candidate is in $\mathcal{C}$ if it is
well-formed), with no reference to a starting point or to how one candidate is
reached from another. $\mathcal{C}$ is countably infinite (node
count, prompt strings, and node-type definitions are all unbounded) yet
strongly regularized: not by a closed library ($\mathcal{N}$ is open) but by the
type discipline its nodes and edges must respect, which is what keeps
$\mathrm{Valid}$ a cheap static check rather than a rollout.

\paragraph{The method-seeded space.}
The open-form $\mathcal{C}$ is far too large to enumerate, and AAS does not try
to. It \emph{seeds} from a published method, ported into a graph as a baseline
$c_0\in\mathcal{C}$, and explores the candidates reachable by applying the
operators above to that seed:
\begin{equation}
  \mathcal{C}(c_0)
  =
  \Big\{\,
    c_0\oplus\Delta
    \ \Big|\
    \Delta\in\mathcal{O}^{*},\;
    \mathrm{Valid}(c_0\oplus\Delta)
  \,\Big\},
  \label{eq:space-operational}
\end{equation}
where $\oplus$ applies the operator patch $\Delta$ to the seed. Because every
operator edits only the candidate's own constituents, the framework runtime and
the vendored environment stay frozen by construction. The search reaches
agents, never the substrate they run on. Every member of $\mathcal{C}(c_0)$ lies
in the open-form $\mathcal{C}$ (it meets the same $\mathrm{Valid}$), so the
seed and the operator budget carve out a tractable region of $\mathcal{C}$ rather
than the whole of it.

\paragraph{An expressive, structurally validatable space.}
Each node is a Python class, hence Turing-complete on its own, and the graph layer
adds loops, branching, and shared state over nodes. Together that is enough that
current embodied agents, workflow-shaped ones in particular, can almost all
be represented cleanly as candidates. What keeps the space searchable is that
validity is a structural, graph-level check: a candidate's wiring (typed edges,
three-pivot coherence, reachable in/out) is screened by the cheap static
$\mathrm{Valid}$ of \eqref{eq:space-declarative}, no rollout, and only authoring a
new node type drops to opaque Python. The space is also not a product of
independent coordinates: an edit to one constituent constrains the others through
$\mathrm{Valid}$, so $\mathcal{C}\neq\prod_k\Pi_k(\mathcal{C})$. The one form it
cannot represent, control flow built dynamically at deployment, is Limitation~1
(\S\ref{sec:limitations}).

\subsection{The search policy: ADAS, AFlow, and KDLoop}
\label{appx:f-optimizer}

The variants share the executable space (\S\ref{appx:f-substrate}) and task
objective (\S\ref{appx:f-eval}). Their search policies specify how proposals
are generated, how memory is retained, and how evaluations are allocated. These choices need not incur the same computational cost. Each variant carries its own run memory $H_t$, a
summary of the evaluated trajectory whose structure is itself part of the method.
The three fall into two shapes: ADAS and AFlow are one family, a single-proposal
archive hill-climb that differs only in how the parent is chosen, while KDLoop takes a
structurally different step.

\subsubsection{ADAS and AFlow: single-proposal archive hill-climb}
\label{appx:f-policy-hillclimb}

Both keep a flat archive $H_t=\{(c_i,f(c_i),\ell_i)\}_{i\le t}$ ($\ell_i$ the
rollout log), draw one parent, and apply the operator set produced by the shared
coding-agent proposer $g$:
\begin{equation}
  c_{t+1}=c_{\mathrm{par}}\oplus g(c_{\mathrm{par}},H_t),
  \qquad c_{\mathrm{par}}\sim P(\cdot\mid H_t).
  \label{eq:policy-hillclimb}
\end{equation}
The generator $g$ (an opaque LLM call emitting $\Delta\in\mathcal{O}^{*}$ as a
native code diff) and the realize\,/\,validate\,/\,evaluate harness around it
(Appendix~\ref{appx:harness}) are identical, and every committed candidate is
measured on the full objective $f$. The two methods differ only in the parent law
$P$:
\[
  \text{ADAS:}\ \ P=\delta_{c_t}\ \text{(archive head)},
  \qquad
  \text{AFlow:}\ \ P=\lambda\,\mathrm{unif}+(1{-}\lambda)\,
  \mathrm{softmax}\!\big(100\,\alpha\,s(c)\big)\ \text{over top-}K,
\]
with $s(c)$ the candidate's bootstrap-median success rate and AFlow's anti-replay
memory steering $g$ off edits already tried on the parent. ADAS edits the last
committed candidate, while AFlow stochastically reselects an archived one.

\subsubsection{KDLoop: a filtered wave over typed memory}
\label{appx:f-policy-kdloop}

KDLoop departs from the single-proposal template. Over a typed memory $H_t$
(facts, open hypotheses, closed-case lessons including refuted edits, and an
intervention-space coverage map), \textsc{Think} emits up to three axis-tagged
operator sets, \textsc{Critic} filters them, and the survivors run as one wave on
the incumbent $c_{\mathrm{par}}$ (the previous iteration's best spec), which
advances greedily to the wave's winner:
\begin{equation}
\begin{gathered}
  \{(a_j,\Delta_j)\}_{j=1}^{m}=\textsc{Think}(H_t)\ \ (m\le3),
  \qquad
  \mathcal{D}=\{\Delta_j:\textsc{Critic}(a_j,\Delta_j,H_t)\},\\[2pt]
  c_{t+1}=\arg\max_{\Delta\in\mathcal{D}}\ \hat f\big(c_{\mathrm{par}}\oplus\Delta\big).
\end{gathered}
  \label{eq:policy-kdloop}
\end{equation}
Here $a_j$ is the intervention axis that the operator set $\Delta_j$ targets, one
lever from the coverage map in $H_t$, and $\mathcal{D}$ collects the proposals
\textsc{Critic} admits. After the wave, \textsc{Distill}/\textsc{Reflect} write
the outcomes back into $H_t$.
The selection signal $\hat f$ is a \textsc{Think}-composed measurement tier,
by default a custom $\ge\!30$-episode subset, escalating to the full objective $f$
only on demonstrated lift, so KDLoop is the one variant that scores wave
candidates more cheaply than the shared evaluator.

\subsection{The evaluation function}
\label{appx:f-eval}

No policy builds the evaluation function $f$. It is absorbed into the AgentCanvas
substrate that already defines the space. Because a candidate is a closed-loop
graph carrying its own environment node (\S\ref{appx:f-substrate}), \emph{running}
it \emph{is} evaluating it: the batch rollout path (\S\ref{appx:ac-eval}) executes
the graph over a fixed search-episode suite and returns the success rate $f(c)$. The same
path serves every variant (KDLoop's cheap signal $\hat f$ is just that path at a
smaller episode budget, the full $f$ using the search-time performance tier), so $f$
is method-free, and the objective $\max_{c\in\mathcal{C}}f(c)$ belongs to the
problem, not to any policy.

Absorbing the evaluator is what makes the representation load-bearing. Unlike
text-domain AAS, where scoring is a cheap deterministic check, here $f$ is an
expensive, noisy multi-episode rollout, so the static $\mathrm{Valid}$ filter of
\eqref{eq:space-declarative} screens malformed candidates \emph{before} any rollout
is spent. And each rollout returns more than a scalar: the logs $\ell$ recorded
alongside $f(c)$ are the evidence a policy's memory $H_t$ distils, so a candidate is
scored with its trace, not as a bare number.

\section{Held-out Evaluation and Significance Test}
\label{app:heldout}

We construct held-out sets by excluding all episodes queried during search
and sampling with seed 42, stratified by scene for navigation and question
type for ExploreEQA. The sets contain 216 MapGPT episodes (72 paths with
three instructions each), 100 SmartWay episodes, and 100 ExploreEQA questions.
Navigation paths are disjoint from those queried during search, but scenes
may overlap. VoxPoser does not require a separate held-out split because
its LIBERO episodes are near-identical.

Frozen baselines and selected candidates receive three evaluation passes
under the search-time model and execution settings. Held-out results do not
feed back into search. We exclude the leak-bearing SmartWay (AFlow)
candidate and ExploreEQA (ADAS), which found no improvement. Add-one
ablations use the same baseline, split, and three-pass protocol.

Table~\ref{tab:heldout-statistics} reports all seven evaluated candidates.
For each episode, we average success over three passes and compute the
difference between candidate and baseline. We use a two-sided paired sign-flip test
and an episode bootstrap, each with 20{,}000 resamples, to obtain $p$-values
and 95\% confidence intervals. Gains are certified when $p<.05$ and the
interval excludes zero, and directional otherwise.

\begin{table}[htbp]
\centering
\small
\setlength{\tabcolsep}{4pt}
\caption{Complete results on held-out episodes. Baseline and candidate SR are
percentages (mean$\pm$sd over three passes). $\Delta$ and its 95\% confidence
interval are in pp. Every $p$-value compares the frozen baseline with the
candidate on the same held-out episodes.}
\label{tab:heldout-statistics}
\begin{tabular}{@{}llrrrcl@{}}
\toprule
Executor & Optimizer & Baseline SR & Candidate SR & $\Delta$ & $p$ & 95\% CI (pp) \\
\midrule
MapGPT & ADAS & $39.4{\pm}2.1$ & $44.4{\pm}1.7$ & $+5.1$ & $.014$ & $[+0.9,+9.3]$ \\
       & AFlow & $39.4{\pm}2.1$ & $46.0{\pm}1.4$ & $+6.6$ & $.007$ & $[+1.9,+11.4]$ \\
       & KDLoop & $39.4{\pm}2.1$ & $44.9{\pm}3.5$ & $+5.6$ & $.016$ & $[+1.1,+10.0]$ \\
\midrule
ExploreEQA & AFlow & $44.3{\pm}0.6$ & $48.3{\pm}0.6$ & $+4.0$ & $.25$ & $[-3.3,+11.3]$ \\
           & KDLoop & $44.3{\pm}0.6$ & $47.7{\pm}1.5$ & $+3.3$ & $.36$ & $[-3.0,+9.7]$ \\
\midrule
SmartWay & ADAS & $27.7{\pm}2.5$ & $18.7{\pm}2.3$ & $-9.0$ & $.005$ & $[-15.0,-3.0]$ \\
         & KDLoop & $27.7{\pm}2.5$ & $28.0{\pm}2.6$ & $+0.3$ & $.98$ & $[-5.3,+6.0]$ \\
\bottomrule
\end{tabular}
\end{table}
\section{Search Replication and Cost}
\label{app:search-robustness}

\subsection{Additional independent searches}
\label{app:search-replication}

We add two independent searches for each of MapGPT (KDLoop), ExploreEQA
(AFlow), and SmartWay (ADAS), using the original model, instructions, and
budget in separate sessions without shared workspaces or memory. Across
three searches per cell, recorded edit inventories overlap by more than
75\%, including 18/20 edits for SmartWay (ADAS), with different discovery orders.

\subsection{Recorded search cost}
\label{app:search-cost}

Table~\ref{tab:search-cost} reports resource usage for the original twelve
searches. Episodes count committed simulator evaluations, and tokens count
orchestrator output. Rollout hours exclude smoke tests and verification
reruns. They sum sequential evaluation durations for ADAS/AFlow and elapsed
evaluation-wave durations for KDLoop, counting concurrent runs once. Missing
KDLoop timings make its rollout hours lower bounds. Orchestration hours are
reported separately. Output tokens and orchestration time were uncapped.

\begin{table}[htbp]
\centering
\small
\setlength{\tabcolsep}{4pt}
\caption{Recorded search usage for the original matrix. Episodes count committed
simulator evaluations. Tokens are orchestrator output tokens (k denotes
thousand and M denotes million). Rollout and orchestration durations are in hours. KDLoop rollout
times are lower bounds because some timing records are missing.}
\label{tab:search-cost}
\begin{tabular}{@{}llrrrr@{}}
\toprule
Executor & Optimizer & Episodes & Output tokens & Rollout h & Orch.\ h \\
\midrule
MapGPT & ADAS & 4{,}536 & 712k & 23.4 & 6.1 \\
       & AFlow & 4{,}536 & 367k & 11.8 & 4.4 \\
       & KDLoop & 3{,}042 & 1.2M & 11.0 & 10.4 \\
\midrule
SmartWay & ADAS & 2{,}100 & 768k & 17.4 & 7.4 \\
         & AFlow & 2{,}000 & 292k & 11.3 & 2.0 \\
         & KDLoop & 1{,}936 & 296k & 10.6 & 4.2 \\
\midrule
ExploreEQA & ADAS & 2{,}100 & 1.0M & 12.5 & 7.3 \\
           & AFlow & 2{,}000 & 267k & 6.0 & 1.8 \\
           & KDLoop & 1{,}628 & 507k & 12.0 & 4.1 \\
\midrule
VoxPoser & ADAS & 4{,}200 & 481k & 6.6 & 3.9 \\
         & AFlow & 4{,}000 & 244k & 6.8 & 1.8 \\
         & KDLoop & 928 & 236k & 2.5 & 3.4 \\
\bottomrule
\end{tabular}
\end{table}

\end{document}